\documentclass[sigconf]{acmart}

\usepackage[utf8]{inputenc} 
\usepackage[T1]{fontenc}    
\usepackage{hyperref}       
\usepackage{url}            
\usepackage{booktabs}       
\usepackage{amsfonts}       
\usepackage{nicefrac}       
\usepackage{microtype}      
\usepackage{xcolor}         

\usepackage{amsmath}
\usepackage{mathtools}
\usepackage{amsthm}
\usepackage{bbm}
\usepackage{cleveref}
\usepackage{bm}
\usepackage{enumitem}
\usepackage{lipsum}
\usepackage{multirow}
\usepackage[normalem]{ulem}
\usepackage{url}

\usepackage{graphicx}
\usepackage{subcaption}
\usepackage{graphicx,color}

\usepackage{wrapfig, flushend, multirow}
\usepackage{longtable}

\usepackage{graphicx}
\usepackage{colortbl}

\usepackage{microtype}
\usepackage{graphicx}
\usepackage{booktabs} 
\usepackage{hyperref}

\newcommand{\name}{ClimateBench-M}

\theoremstyle{plain}
\newtheorem{theorem}{Theorem}[section]

\newtheorem{lemma}[theorem]{Lemma}

\theoremstyle{definition}

\theoremstyle{remark}
\newtheorem{remark}[theorem]{Remark}

\AtBeginDocument{%
  }

\setcopyright{acmlicensed}
\copyrightyear{2025}
\acmYear{2025}
\acmDOI{XXXXXXX.XXXXXXX}
\acmConference[KDD '25]{the 31st ACM SIGKDD Conference on Knowledge Discovery and Data Mining}{August 3--7, 2025}{Toronto, Canada}
\acmISBN{978-1-4503-XXXX-X/18/06}

\begin{document}

\title{\name: A Multi-Modal Climate Data Benchmark\\ with a Simple Generative Method}

\author{Dongqi Fu, Yada Zhu, Zhining Liu, Lecheng Zheng, Xiao Lin, Zihao Li, Liri Fang, Katherine Tieu, Onkar Bhardwaj, Kommy Weldemariam, Hanghang Tong, Hendrik Hamann, Jingrui He}
\affiliation{
  \institution{University of Illinois Urbana-Champaign, IBM Research \\
  \{dongqifu, liu326, lecheng4, xiaol13, zihaoli5, lirif2, kt42, htong, jingrui\}@illinois.edu\\
  \{yzhu, onkarbhardwaj, kommy, hendrikh\}@ibm.us.com
  }
  \country{}
}

\renewcommand{\shortauthors}{Fu et al.}

\begin{abstract}
Climate science studies the structure and dynamics of Earth’s climate system and seeks to understand how climate changes over time, where the data is usually stored in the format of time series, recording the climate features, geolocation, time attributes, etc. 
Recently, much research attention has been paid to the climate benchmarks. In addition to the most common task of weather forecasting, several pioneering benchmark works are proposed for extending the modality, such as domain-specific applications like tropical cyclone intensity prediction and flash flood damage estimation, or climate statement and confidence level in the format of natural language.
To further motivate the artificial general intelligence development for climate science, in this paper, we first contribute a multi-modal climate benchmark, i.e., \textbf{\name}, which aligns (1) the time series climate data from ERA5, (2) extreme weather events data from NOAA, and (3) satellite image data from NASA HLS based on a unified spatial-temporal granularity. Second, under each data modality, we also propose a simple but strong generative method that could produce competitive performance in weather forecasting, thunderstorm alerts, and crop segmentation tasks in the proposed \name. The data and code of \name\ are publicly available at \url{https://github.com/iDEA-iSAIL-Lab-UIUC/ClimateBench-M}.
\end{abstract}




\maketitle

\section{Introduction}
Climate science investigates the structure and dynamics of earth’s climate system and seeks to understand how global, regional, and local climates are maintained as well as the processes by which they change over time,\footnote{\url{https://plato.stanford.edu/entries/climate-science/}}
In general, climate data is usually represented by a time series numerical format that covers climate features (e.g., temperature, wind, and atmospheric water content), geolocation information (e.g., longitude, latitude, and geocode), and time (e.g., hours and days).
Recently, to develop artificial intelligence techniques for climate science, many interesting climate benchmarks have been proposed.

For example, \textit{WeatherBench}~\cite{rasp2020weatherbench} provides a common data set and evaluation metrics to enable direct comparison between different data-driven approaches to medium-range weather forecasting (3-5 days lead time). Stephan et al.~\cite{rasp2020weatherbench} argue that the traditional weather models based on physical equations have limitations, and data-driven approaches like deep learning could potentially produce better forecasts by learning directly from observations. The data set includes preprocessed ERA5, and the paper provides baseline results using linear regression, deep learning, and physical models. Following that, \textit{WeatherBench 2}~\cite{DBLP:journals/corr/abs-2308-15560}
aims to accelerate progress in data-driven weather modeling by providing an open-source evaluation framework, publicly available training and ground truth data, and a continuously updated website with the latest metrics and state-of-the-art models. The benchmark is designed to closely follow the forecast verification practices used by operational weather centers, with a set of headline scores to provide an overview of model performance.

\begin{figure*}[t]
\scalebox{0.97}{
\begin{tabular}{cc}
\includegraphics[width=0.58\textwidth]{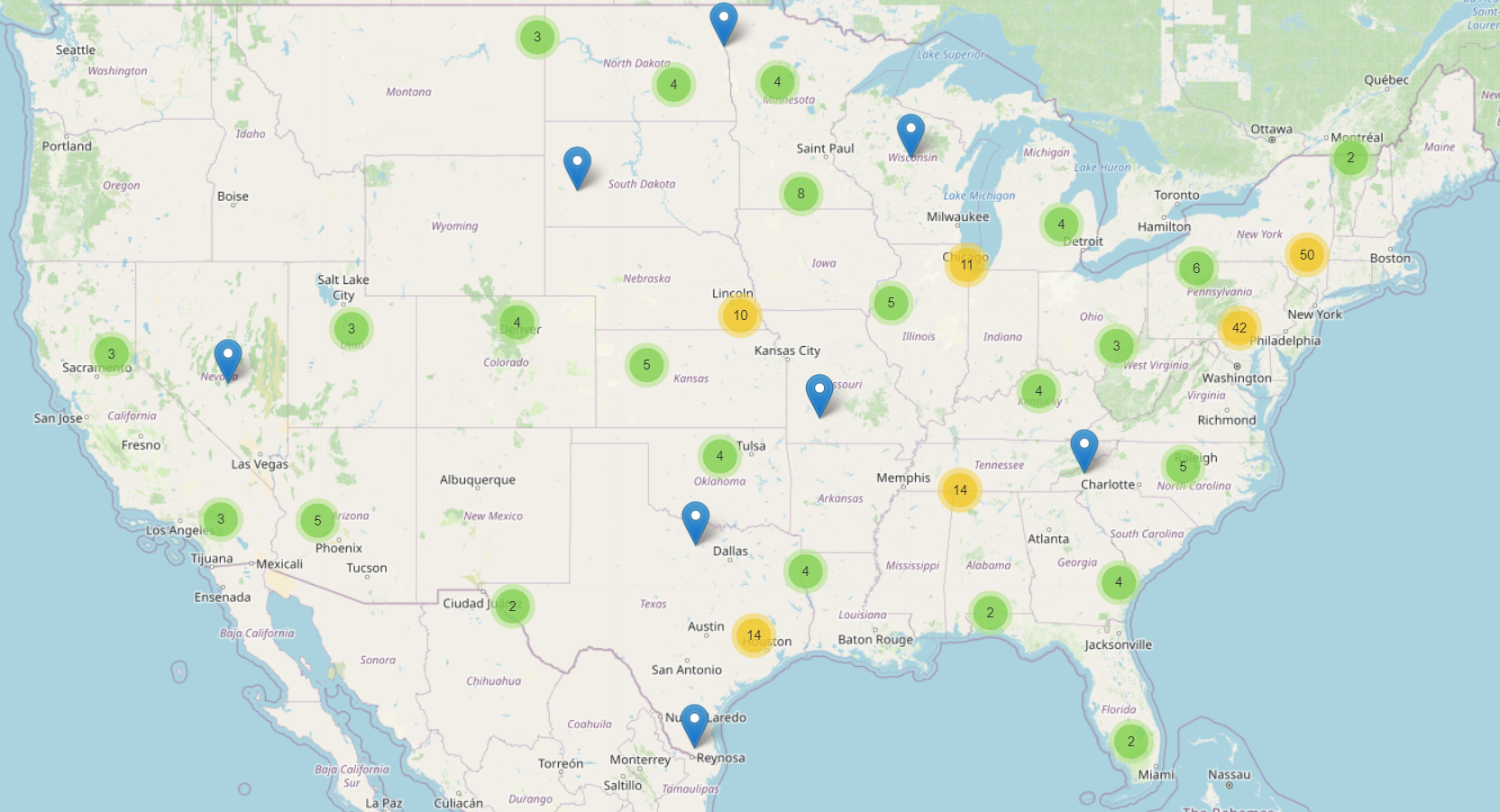} & \includegraphics[width=0.41\textwidth]{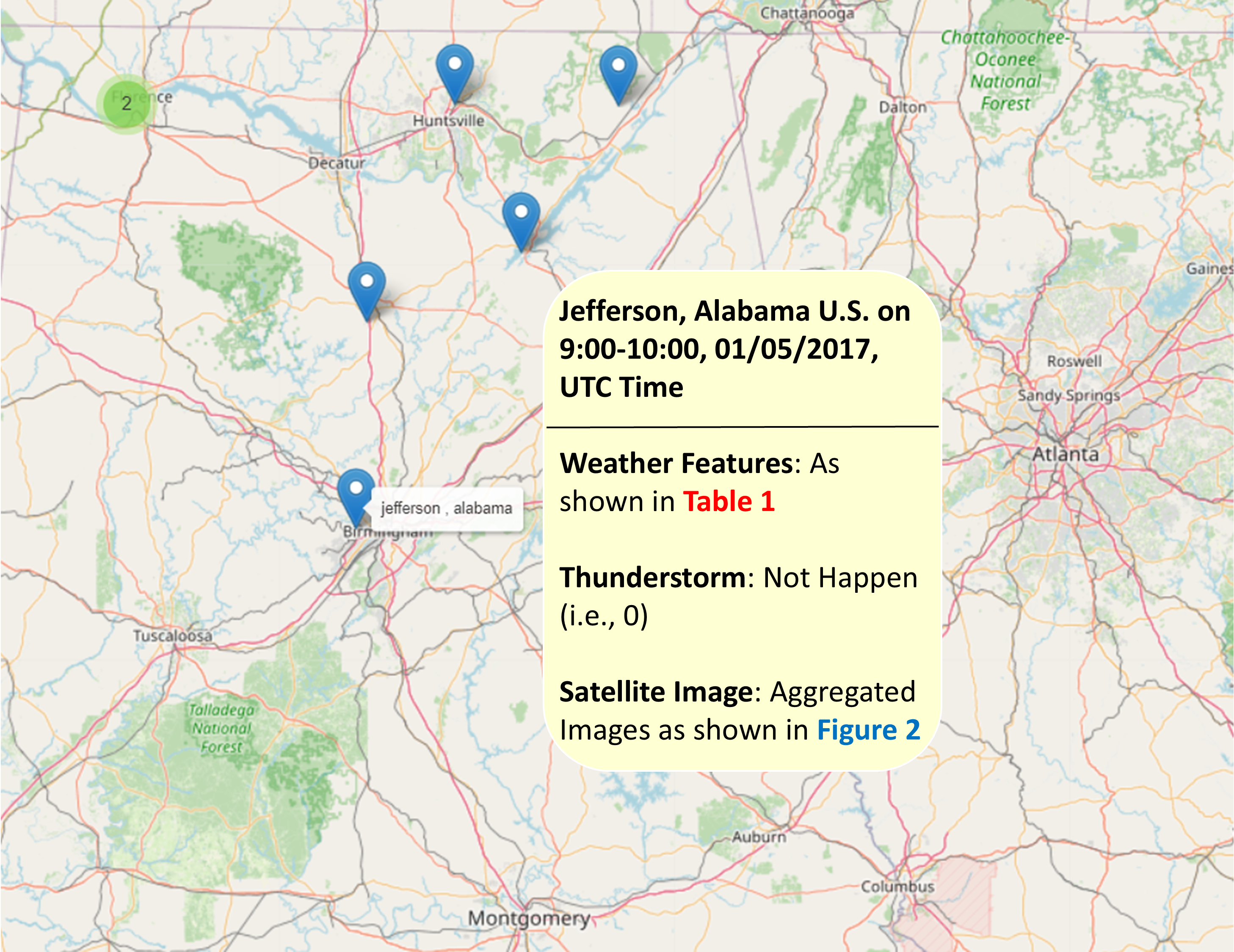}
\end{tabular}}
\centering
\caption{\textbf{Left:} Geographic Distribution of Covered Counties in \name\ (The number in the circle stands for the aggregation of nearby counties) \textbf{Right:} A Specific Example of Jefferson, Alabama U.S. on 9:00-10:00, 01/05/2017, UTC Time}
\label{Fig:geo_vis}
\end{figure*}

\begin{table*}[h]
\caption{(Part of) Feature Descriptions with Instance Values Sampled from Jefferson, Alabama U.S. on 9:00-10:00, 01/05/2017, UTC. Full features are in Table~\ref{tab:features} in Appendix~\ref{weather_feature_description}.}
\begin{tabular}{p{2.5cm}p{2cm}p{7cm}p{2.5cm}}
    \toprule
    Feature    & Unit & Description & Value\\
    \midrule
    100-meter wind towards east  &m s$^{-1}$ &This parameter is the eastward component of the 100 m wind. It is the horizontal speed of air moving towards the east, at a height of 100 meters above the surface of the Earth, in meters per second. Care should be taken when comparing model parameters with observations, because observations are often local to a particular point in space and time, rather than representing averages over a model grid box. This parameter can be combined with the northward component to give the speed and direction of the horizontal 100 m wind. & -3.192476\\
    \midrule
    100-meter wind towards north  &m s$^{-1}$ &This parameter is the northward component of the 100 m wind. It is the horizontal speed of air moving towards the north, at a height of 100 meters above the surface of the Earth, in meters per second. Care should be taken when comparing model parameters with observations, because observations are often local to a particular point in space and time, rather than representing averages over a model grid box. This parameter can be combined with the eastward component to give the speed and direction of the horizontal 100 m wind. & -1.892055\\
    \midrule
    10-meter wind gust (maximum)  &m s$^{-1}$ &Maximum 3-second wind at 10 m height as defined by WMO. Parametrization represents turbulence only before 01102008; thereafter effects of convection are included. The 3 s gust is computed every time step, and the maximum is kept since the last postprocessing. & 3.620435\\
    \midrule
    Atmospheric water content   & kg m$^{-2}$ & This parameter is the sum of water vapor, liquid water, cloud ice, rain, and snow in a column extending from the surface of the Earth to the top of the atmosphere. In old versions of the ECMWF model (IFS), rain and snow were not accounted for. & 9.287734\\
    \bottomrule
    \label{tab:part_feature}
\end{tabular}
\end{table*}
In addition to the weather forecasting climate benchmarks, some task-specific and domain-specific benchmarks are proposed. For example, the authors in~\cite{DBLP:conf/nips/RacahBMKPP17} present a large-scale climate dataset called \textit{ExtremeWeather}, which is designed to encourage machine learning research in the detection, localization, and understanding of extreme weather events, to further address the problem that the existing labeled data for climate patterns like hurricanes, extra-tropical cyclones, and weather fronts can be incomplete.
Also, \textit{FloodNet}~\cite{DBLP:journals/access/RahnemoonfarCSV21} presents a high-resolution aerial imagery dataset, which was captured after Hurricane Harvey to aid in post-flood scene understanding to alleviate the problem that the existing natural disaster datasets are limited, with satellite imagery having low spatial resolution and ground-level imagery from social media being noisy and not scalable.
With the success of large language models (LLMs)~\cite{DBLP:journals/corr/abs-2303-18223}, 
\textit{ClimateX}~\cite{DBLP:journals/corr/abs-2311-17107} presents a novel, curated, expert-labeled dataset of 8,094 climate statements from the latest IPCC reports, labeled with their associated confidence levels. The authors use this dataset to evaluate how accurately recent LLMs can classify human expert confidence in climate-related statements.

Those aforementioned benchmarks pave the way for developing possible artificial intelligence techniques for climate science from one single aspect.
Then, a natural question arises: \textbf{can we provide a comprehensive climate benchmark that has multiple data modalities for chasing the artificial general intelligence~\cite{DBLP:journals/corr/abs-2303-12712} (AGI)} for climate applications?
To speed up the AGI development for climate science, in this paper, we first propose a multi-modal climate benchmark named \name, which aligns the ERA5~\cite{hersbach2018era5}\footnote{\url{https://cds.climate.copernicus.eu/cdsapp\#!/home}} time series data for weather forecasting, NOAA~\footnote{\url{https://www.ncdc.noaa.gov/stormevents/ftp.jsp}} extreme weather events records for extreme weather alerts, and HLS~\cite{HLS_Foundation_2023}~\footnote{\url{https://huggingface.co/datasets/ibm-nasa-geospatial/multi-temporal-crop-classification}} satellite image data for the crop segmentation, based on a unified spatial-temporal granularity. Moreover, we also propose a simple generative model, called SGM, for each task in the proposed \name. SGM is based on the encoder-decoder framework, and the choices of encoders and decoders vary for different tasks. Overall, in each task of \name, SGM produces a competitive performance with different baseline methods.


\section{\name}

\subsection{Datasets}
\name\ benchmark aligns three datasets from different modalities based on the spatial and temporal granularity.
The raw data originates from public datasets \textbf{ERA5}~\cite{hersbach2018era5}\footnote{\url{https://cds.climate.copernicus.eu/cdsapp\#!/home}}, \textbf{NOAA}~\footnote{\url{https://www.ncdc.noaa.gov/stormevents/ftp.jsp}} and \textbf{NASA HLS}~\cite{HLS_Foundation_2023}~\footnote{\url{https://huggingface.co/datasets/ibm-nasa-geospatial/multi-temporal-crop-classification}}.
\begin{itemize}
    \item ERA5 provides hourly estimates for a large number of atmospheric, ocean-wave and land-surface quantities. The data is available from 1940 onwards.
    \item NOAA is National Oceanic and Atmospheric Administration that has the National Centers for Environmental Information (NCEI), which center published the Storm Events Database, currently recording the data from January 1950 to February 2024, as entered by NOAA's National Weather Service (NWS).
    \item The NASA HLS (Harmonized Landsat and Sentinel-2) v2.0 dataset integrates high-resolution, multi-spectral satellite images from Landsat and Sentinel-2 missions, spanning from 2013 to the present. 
\end{itemize}

\begin{figure*}[h]
  \centering
  \includegraphics[width=1\textwidth]{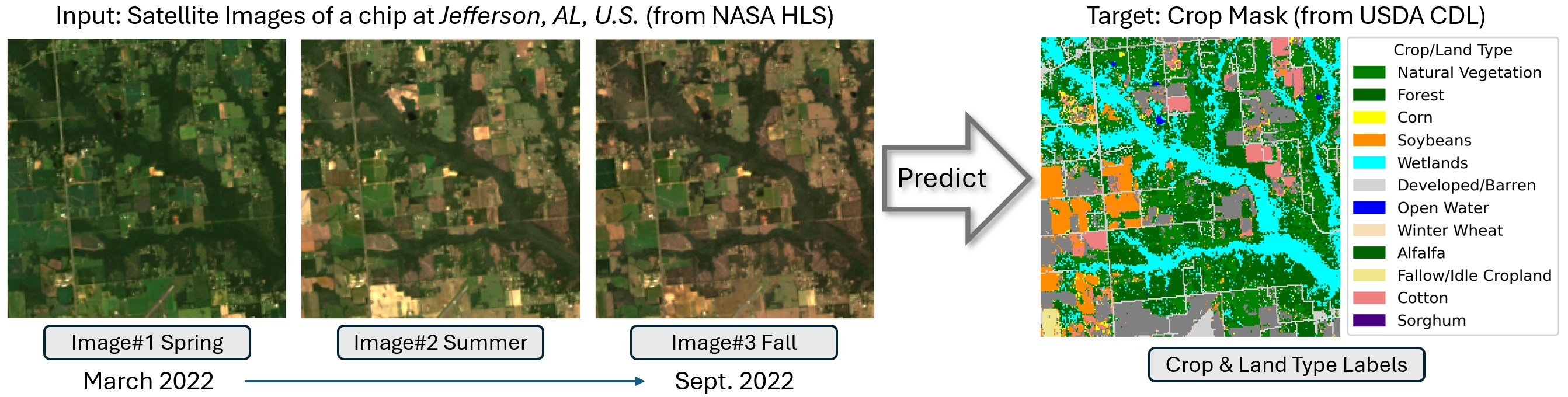}
  \caption{Example of the crop type segmentation task based on NASA HLS and USDA CDL.}
  \label{Fig:cropseg}
\end{figure*}

\subsection{Data Preprocessing and Alignment}
First, NOAA is a thunderstorm dataset, which has a minute-level record denoting whether the thunderstorm happens or not in this minute. The location is marked by the county name and FIPS geocode (e.g., Jefferson, 73) and state name and FIPS geocode (Alabama, 1).
Therefore, with the support and knowledge of our domain experts, we selected 45 thunderstorm-related weather features from ERA5 (e.g., wind gusts, rain, etc.) of 238 counties in the United States of America from 2017 to 2020. The details of these 45 weather features are specified in Table~\ref{tab:features} in Appendix~\ref{weather_feature_description}.

The geographic distribution of 238 selected counties in the United States of America is shown in Figure~\ref{Fig:geo_vis}, where the circle with numbers denotes the aggregation of spatially near counties. The left part of Figure~\ref{Fig:geo_vis} shows the detailed information of Jefferson, Alabama U.S. on 9:00-10:00, 01/05/2017, UTC Time, with the corresponding weather feature in Table~\ref{tab:part_feature} and satellite image in Figure~\ref{Fig:cropseg}.

To be specific, among the 238 selected counties, 100 are selected for the top-ranked counties based on the yearly frequency of thunderstorms. The rest are selected randomly to try to provide extra information (e.g., causal effect). Because we chose thunderstorms as the anomaly pattern to be detected after forecasting, we then mapped the name and code of locations in the NOAA dataset with the latitude and longitude of locations in the ERA5 dataset. After that, for each specific county, each row (i.e., hour) of 45 weather features in ERA5 will be associated with a thunderstorm label; if any minute in this hour has the thunderstorm record, then 1 will be marked; otherwise, 0 will be marked. The spatial-temporal distribution of thunderstorms in \name\ is shown in Table~\ref{tb:label_distribution}.

\begin{table}[h]
\caption{Statistics of Thunderstorm Records in \name\ over 238 Selected Counties in the United States from 2017 to 2021}
\centering
\scalebox{1}{
\begin{tabular}{cccccc}
\hline
Year & 2017 & 2018 & 2019 & 2020 & 2021 \\ \hline
Jan  & 26   & 3    & 2    & 41   & 7    \\ 
Feb  & 53   & 6    & 9    & 50   & 8    \\ 
Mar  & 85   & 16   & 26   & 63   & 62   \\ 
Apr  & 93   & 44   & 140  & 170  & 60   \\ 
May  & 245  & 207  & 263  & 175  & 218  \\ 
Jun  & 770  & 302  & 348  & 331  & 452  \\ 
Jul  & 306  & 291  & 457  & 453  & 701  \\ 
Aug  & 294  & 269  & 415  & 354  & 435  \\ 
Sep  & 61   & 80   & 122  & 29   & 123  \\ 
Oct  & 32   & 32   & 82   & 60   & 55   \\ 
Nov  & 20   & 22   & 9    & 114  & 11   \\ 
Dec  & 5    & 15   & 11   & 8    & 58   \\ \hline
\end{tabular}}
\label{tb:label_distribution}
\end{table}

\begin{figure*}[t]
\includegraphics[width=1\textwidth]{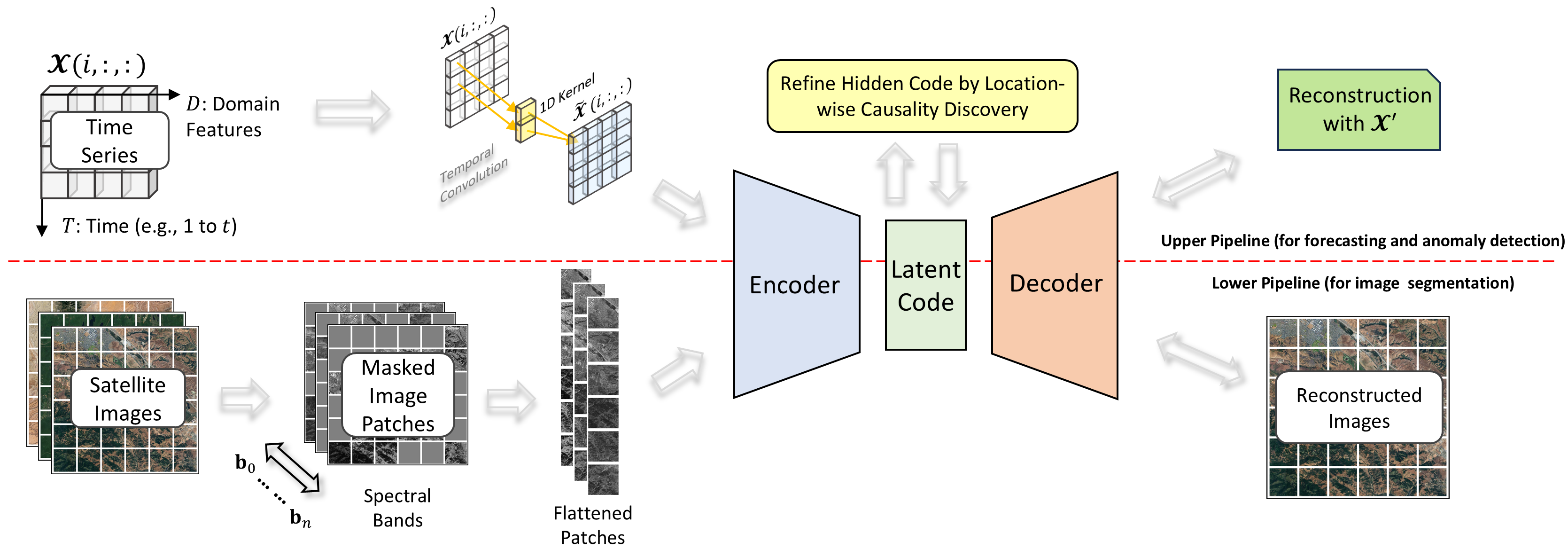}
\centering
\caption{The Proposed Simple Generative Model (SGM). The upper level of the figure shows the time series forecasting pipeline, and the lower level of the figure shows the image segmentation pipeline. Two pipelines have different choices of encoders and decoders.}
\label{Fig:new_framework}
\end{figure*}

For NASA HLS satellite image dataset, we create a crop segmentation and classification task by deriving pixel-level labels from USDA's Crop Data Layer (CDL). 
\begin{itemize}
    \item First, a set of 5,000 chips was defined based on samples from the USDA CDL to ensure a representative sampling across the continental United States. 
    \item We then spatially align these chips with the 238 counties contained in the ERA5 data based on latitude and longitude.
    \item Specifically, for each chip-county pair, we check the average difference in latitude and longitude between the center point of the chip and the county. If the difference is less than 1, we assign the chip to the corresponding county. If multiple counties meet the criteria, we assign the chip to the nearest county to ensure no overlap between chips within each county, thus preventing data leakage when performing county-based train/test split.
    \item For each chip, we retrieve 3 satellite images from the NASA HLS dataset evenly distributed in time from March to September 2022 to capture the ground view at different stages of the season.
    \item Finally, we perform an image quality check on each chip using the metadata, discarding any chip with clouds, cloud shadows, or missing values.
\end{itemize}
After all matching and filtering, we obtain 3138 valid chips corresponding to 169 counties.
For each chip, the input GeoTIFF image file covers a 224 x 224 pixel area at 30m spatial resolution with 18 spectral bands (6 spectral bands of 3 images stacked together).
The predicted target is a same-size GeoTIFF file with a single band recording the target class for each pixel.

\subsection{Task 1: Weather Forecasting}

\textbf{Notations}. We denote the weather time series data stored in $\mathcal{X} \in \mathbb{R}^{N \times D \times T }$. Note that a slice of $\mathcal{X}$, i.e., $\mathcal{X}(i, :, :) \in \mathbb{R}^{D \times T }$, $i \in \{1, \ldots, N\}$, is typically denoted as the common multivariate time series data~\cite{DBLP:conf/kdd/SuZNLSP19, DBLP:conf/icdm/ZhaoWDHCTXBTZ20}.
For example, in each element $\mathcal{X}(i,d,t)$ of the nationwide weather data $\mathcal{X}$, $i \in \{1 \ldots, N\}$ can be the number of spatial locations (e.g., counties), $d \in \{1 \ldots, D\}$ can be the dimension of weather features (e.g., temperature and humidity), and $t \in \{1 \ldots, T\}$ can be the timestamp (e.g., hour). Throughout the paper, we use the calligraphic letter to denote a 3D tensor (e.g., $\mathcal{X}$) and the bold capital letter to denote a 2D matrix (e.g., $\bm{X}$). 

\textbf{Problem Definition}. Given the time series data $\mathcal{X} \in \mathbb{R}^{N \times D \times T}$, we aim to forecast the future data $\mathcal{X}' \in \mathbb{R}^{N \times D \times \tau}$, where $\tau$ is a forecasting window.

\subsection{Task 2: Thunderstorm Alerts}
\textbf{Problem Definition}. Recall that, in the forecasting task, we aim to forecast the future data $\mathcal{X}' \in \mathbb{R}^{N \times D \times \tau}$ from the history data $\mathcal{X} \in \mathbb{R}^{N \times D \times T}$. For achieving the thunderstorm alert task, we also aim to find the anomaly in the forecast, i.e., with the forecast $\mathcal{X}'$, we aim to detect if $\mathcal{X}'$ contains abnormal values, i.e., whether thunderstorms happens in a certain location on a certain future hour based on the forecasting window.

\subsection{Task 3: Crop Segmentation}

\textbf{Notations}.  
For the crop segmentation task, we collect a series of satellite images at different times but at the same place, aiming to distinguish the crop types in various regions within those images. Specifically, we denote the satellite images as $\mathcal{X} \in \mathbb{R}^{N\times D \times T}$, where $N$ represents the number of pixels within the images, $D$ represents the number of channels (e.g., RGB brand, near-infrared, and shortwave infrared), and $T$ represents the number of images at the same place. We also denote the crop types as $\mathbf{y} \in \mathbb{R}^N$, and $\mathbf{y}(i), i \in \{1, \dots, N\}$ represents the type of crop grown in the area corresponding to the $i$-th pixel.

\textbf{Problem Definition}. Given the image data $\mathcal{X} \in \mathbb{R}^{N \times D \times T}$, we aim to predict the crop type of each pixel $\mathbf{y} \in \mathbb{R}^N$, as shown in Figure~\ref{Fig:cropseg}.

\section{Simple Generative Model (SGM)}
In this section, we first give an overview of SGM and then induce the details of applying it to different tasks of \name\ benchmark.

\subsection{Overview}
As shown in Figure~\ref{Fig:new_framework}, the SGM is based on an encoder-decoder framework and has two pipelines. The upper pipeline is for time series forecasting (targeting the weather forecasting task) and anomaly detection (targeting the thunderstorm alerts). The lower pipeline is for image segmentation (targeting the temporal crop segmentation).

\subsection{Deployment of SGM for Time Series Forecasting and Anomaly Detection}

In this section, we briefly introduce how the upper pipeline of SGM achieves time series forecasting and anomaly detection. The detailed information can also be found in our previous paper~\cite{DBLP:journals/corr/abs-2408-04254}.

In addition to forecasting, the upper pipeline of SGM is also responsible for anomaly detection. Thus, we design the hidden feature $\mathcal{H}$ extraction in the upper pipeline of SGM motivated by the Extreme Value Theory~\cite{beirlant2004statistics} or so-called Extreme Value Distribution in stream~\cite{ DBLP:conf/kdd/SifferFTL17}.

\begin{remark}
According to the Extreme Value Distribution~\cite{fisher1928limiting}, under the limiting forms of frequency distributions, extreme values have the same kind of distribution, regardless of original distributions.
\label{remark}
\end{remark}

An example~\cite{DBLP:conf/kdd/SifferFTL17} can help interpret and understand the Extreme Value Distribution theory. Maximum temperatures or tide heights have more or less the same distribution even though the distributions of temperatures and tide heights are not likely to be the same. As rare events have a lower probability, there are only a few possible shapes for a general distribution to fit.

Inspired by this observation, we can design a simple but effective module in SGM to achieve anomaly detection along with the forecasting, i.e., an encoder-decoder model that tries to explore the distribution of normal features in $\mathcal{X}$ as shown in Figure~\ref{Fig:new_framework}. As long as this encoder-decoder model can capture the latent distribution for normal events, then the generation probability of a piece of time series data can be utilized as the condition for detecting anomaly patterns. This is because the extreme values are identified with a remarkably low generation probability. To be specific, after the forecast $\bm{H}^{(t)}$ is output, the generation probability of $\bm{H}^{(t)}$ into $\bm{X}^{(t)}$ can be used as the evidence to detect the anomalies at $t$. The transformation from $\bm{X}^{(t)}$ to $\bm{H}^{(t)}$ can be realized by a model-agnostic pre-trained autoencoder.

Moreover, we use the mean absolute error (MAE) loss on the prediction and the ground truth, which is effective and widely applied to time-series forecasting tasks~\cite{DBLP:conf/iclr/LiYS018, DBLP:conf/iclr/Shang0B21}.
\begin{equation}
   \min_{\Theta_{i}, \bm{A}^{(t-1)}, \ldots, \bm{A}^{(t-l)} } \mathcal{L}_{pred} = \sum_{i} \sum_{t} | H(i,:)^{(t)} - \hat{H}(i,:)^{(t)}|
\label{eq: outer_opt}
\end{equation}
where $\Theta_{i}, \bm{A}^{(t-1)}, \ldots, \bm{A}^{(t-l)}$ are all learnable parameters for the prediction $\hat{H}(i,:)^{(t)}$ of variable $i$ at time $t$.
Note that $\bm{A}^{(t-1)}$ is a learnable parameter denoting the causal effects among all locations at time $t$ for better forecasting performance (as shown in Figure~\ref{Fig:dag_vis}), and the learning simply relies on the Structural Equation Model (SEM)\cite{DBLP:conf/nips/ZhengARX18}, and the details are shown in the Appendix~\ref{sec:causality}.

To be more specific, $f_{\Theta_{i}}$ is a sequence-to-sequence model~\cite{DBLP:conf/nips/SutskeverVL14}, which means that given a time window (or time lag), SGM could forecast the corresponding features for the next time window.
\begin{equation}
\label{eq:forecast}
    \hat{H}(i,:)^{(t)} =  f_{\Theta_{i}}[(\bm{A}^{(t-1)}, \bm{H}^{(t-1)}), \ldots, (\bm{A}^{(t-L)}, \bm{H}^{(t-L)})]
\end{equation}
where $L$ is the lag (or window size) in the Granger Causality, and $i$ is the index of the $i$-th variable. $f_{\Theta_{i}}$ is a neural computation unit with all parameters denoted as $\Theta_{i}$, whose input is an $L$-length time-ordered sequence of $(\bm{A}, \bm{H})$. And $f_{\Theta_{i}}$ is responsible for estimating variable $i$ at time $t$ from all variables that occurred in the past time lag $l$. In the upper pipeline of the proposed SGM model, we use graph recurrent neural networks~\cite{DBLP:journals/tnn/WuPCLZY21}.

\subsection{Deployment of SGM for Image Segmentation}
In the task of crop classification, we use mmsegmentation~\cite{mmseg2020}, an OpenMMLab Semantic Segmentation Toolbox, to segment the satellite images, following \cite{jakubik2023foundation}.

To handle the crop satellite image, we choose vision transformer~\cite{DBLP:conf/iclr/DosovitskiyB0WZ21} as the backbone of the encoder-decoder pairs for our proposed SGM. We use random crop and random flip to augment the training data. 

\begin{table*}[ht]
\caption{Forecasting Error (MAE, $10^{-2}$)}
\centering
\scalebox{1}{
\begin{tabular}{ccccc}
\hline
      & ERA5-2017 $(\downarrow)$          & ERA5-2018 $(\downarrow)$          & ERA5-2019 $(\downarrow)$          & ERA5-2020 $(\downarrow)$ \\ \hline
GRU   & 1.8834 $\pm$ 0.0126               & 1.9764 $\pm$ 0.1466               & 1.6194 $\pm$ 0.2645               & 1.7859 $\pm$ 0.2324           \\
DCRNN & 0.0819 $\pm$ 0.0025               & 0.0797 $\pm$ 0.0049               & 0.0799 $\pm$ 0.0035               & 0.0826 $\pm$ 0.0033           \\
GTS   & 0.0777 $\pm$ 0.0054               & 0.0766 $\pm$ 0.0029               & 0.0760 $\pm$ 0.0031               & 0.0742 $\pm$ 0.0021           \\
SGM & 0.0496 $\pm$ 0.0017              & 0.0499 $\pm$ 0.0017               & 0.0502 $\pm$ 0.0016               & 0.0488 $\pm$ 0.0019 \\
ST-SSL & 0.0345 $\pm$ 0.0051               & 0.0330 $\pm$ 0.0018               & 0.0361 $\pm$ 0.0021               & 0.0348 $\pm$ 0.0020           \\
SGM++ & $\textbf{0.0271}$ $\pm$ $\textbf{0.0004}$ & $\textbf{0.0276}$ $\pm$ $\textbf{0.0004}$ & $\textbf{0.0282}$ $\pm$ $\textbf{0.0003}$ & $\textbf{0.0265}$ $\pm$ $\textbf{0.0004}$ \\\hline
\end{tabular}}
\label{tb:forecasting}
\end{table*}

\section{Experiments}
In this section, we report the performance of our SGM and different baseline methods in each task of \name.

\subsection{Evaluation Metrics}
We measure the performance of the baseline methods as well as the proposed method on the \name\ with respect to the following metrics:

(1) \textbf{Accuracy (Acc)}: It evaluates the overlap between the prediction and the ground-truth, i.e., $\text{Acc} = \frac{a}{b}$, where $a$ is the number of correct prediction and $b$ is the total number of samples. 

(2) \textbf{Mean Absolute Error (MAE)}: It assess the difference between the prediction and the ground truth, which is defined in Equation~\ref{eq: outer_opt}.

(3) \textbf{Intersection of Union (IoU)}: It measures the ratio of the intersection of two sets over the union of two sets as follows:
\begin{equation}
    \text{IoU}=\frac{|A \cap B|}{|A \cup B|}
\end{equation}
where $A$ and $B$ are the prediction set and the ground-truth set, respectively. 

(4) \textbf{Area Under the Receiver Operating Characteristic Curve (AUC-ROC)}: 
It quantifies the ability of a model to distinguish between classes by measuring the area under the ROC curve. The ROC curve is a plot of the true positive rate (TPR) against the false positive rate (FPR) at various threshold settings. AUC-ROC is defined as:
\begin{equation}
    \text{AUC-ROC} = \int_0^1 b(a) ~da
\end{equation}
where $a$ and $b$ are TPR and FPR, respectively.

\subsection{Baselines}
The first category is for tensor time series forecasting: (1) GRU~\cite{DBLP:journals/corr/ChungGCB14} is a classical sequence to sequence generative model. (2) DCRNN~\cite{DBLP:conf/iclr/LiYS018} is a graph convolutional recurrent neural network, of which the input graph structure is given and shared by all timestamps. To obtain that graph, we let each node randomly distribute it's unit weights as the probability of connecting other nodes. (3) GTS~\cite{DBLP:conf/iclr/Shang0B21} is also a graph convolutional recurrent neural network that does not need the input graph but learns the structure based on the node features, but the learned structure is also shared by all timestamps and is not causal. To compare the performance of DCGNN~\cite{DBLP:conf/iclr/LiYS018} and GTS~\cite{DBLP:conf/iclr/Shang0B21} with SGM, causality is the control variable since we make all the rest (e.g., neural network type, number of layers, etc.) identical.

The second category is for anomaly detection on tensor time series: (1) DeepSAD~\cite{DBLP:conf/iclr/RuffVGBMMK20}, (2) DeepSVDD~\cite{DBLP:conf/icml/RuffGDSVBMK18}, and (3) DROCC~\cite{DBLP:conf/icml/GoyalRJS020}. Since these three methods have no forecast abilities, we let them use the ground-truth observations, and our SGM utilizes the forecast features during anomaly detection experiments. Also, these three baselines are designed for multi-variate time-series data, not tensor time-series. Thus, we flatten our tensor time series along the spatial dimension and report the average performance for these three baselines over all locations.

The third category is for image segmentation: (1) DeepLabV3 \cite{chen2017rethinking} is a semantic image segmentation model that utilizes atrous convolution to adjust filter field-of-view and capture multi-scale context with multiple atrous rates. (2) Swin \cite{liu2021swin} is a hierarchical vision Transformer model that uses a shifted windowing scheme to efficiently handle segmentation tasks by computing self-attention within non-overlapping local windows. Since the aforementioned baselines do not inherently incorporate temporal dependencies, we concatenate all images at the same location along the channel dimension and utilize the combined image for segmentation.

\subsection{Forecasting}
In Table~\ref{tb:forecasting}, we present the forecasting performance in terms of mean absolute error (MAE) on the testing data of three algorithms, namely DCGNN~\cite{DBLP:conf/iclr/LiYS018}, GTS~\cite{DBLP:conf/iclr/Shang0B21}, ST-SSL~\cite{DBLP:conf/aaai/JiW00XWZZ23}, our SGM, and SGM++ (i.e., SGM with persistence forecast constraints). Here, we set the time window as 24, meaning that we use the past 24 hours tensor time series to forecast the future 24 hours in an autoregressive manner.
Moreover, for baselines and SGM, we set $f_{\Theta_{i}}$ in Eq.\ref{eq:forecast} shared by all weather variables to ensure the scalability, such that we do not need to train $N$ recurrent graph neural networks for a single prediction.

In Table~\ref{tb:forecasting}, we can observe a general pattern that our SGM outperforms the baselines with GTS performing better than DCGNN. For example, with 2017 as the testing data, our SGM performs 39.44\% and 36.16\% better than DCRNN and GTS.
An explanation is that the temporally fine-grained causal relationships can contribute more to the forecasting accuracy than non-causal directed graphs, since DCGNN, GTS, and our SGM all share the graph recurrent manner. SGM, however, discovers causalities at different timestamps, while DCGNN and GTS use feature-similarity-based connections. Moreover, ST-SSL achieves competitive forecasting performance via contrastive learning on time series. Motivated by a contrastive manner, SGM++ is proposed by persistence forecast constraints. That is, the current forecast of SGM is further calibrated by its nearest time window (i.e., the last 24 hours in our setting). The detailed implementation is provided in Appendix~\ref{sec:implementation}.

\begin{figure}[h]
\includegraphics[width=0.49\textwidth]{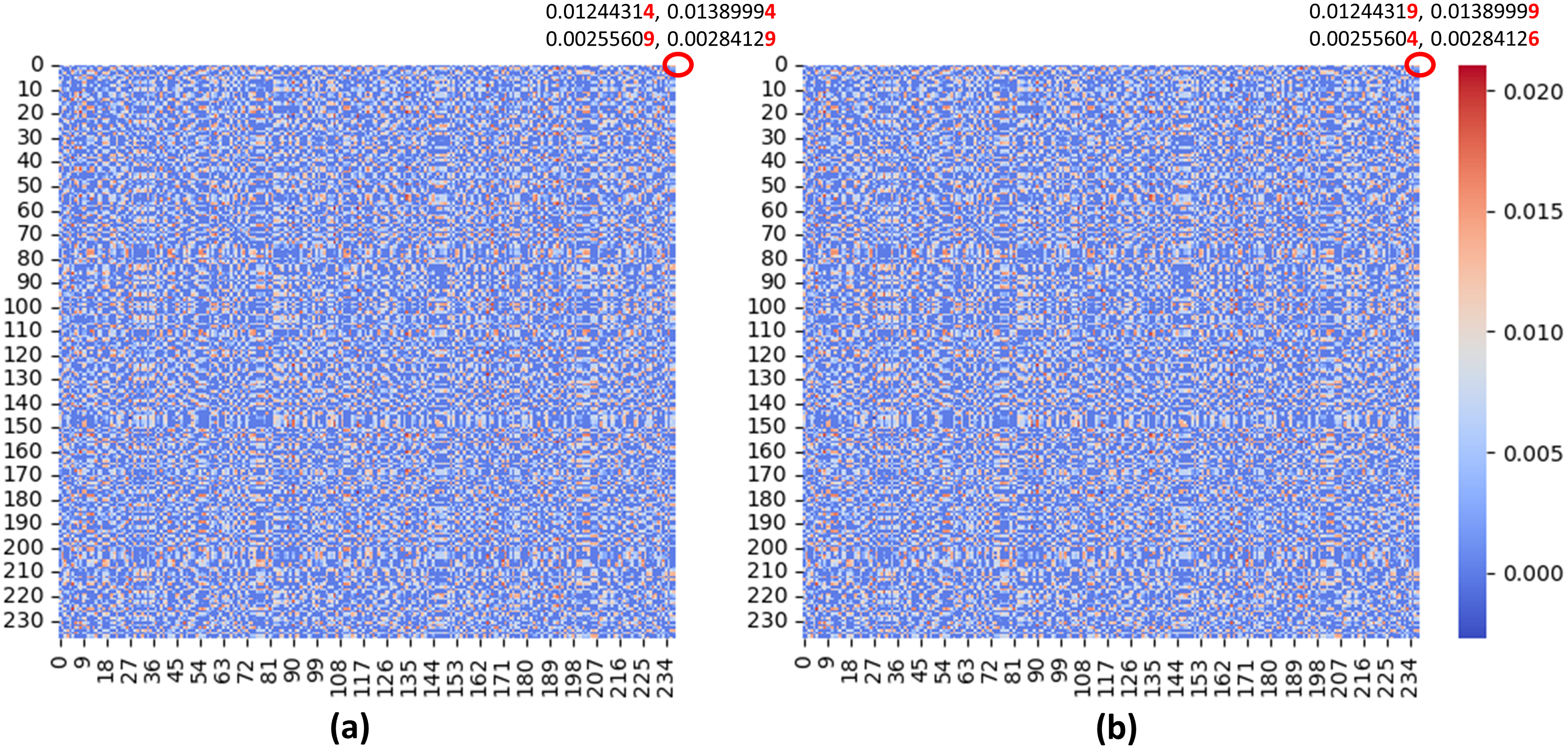}
\centering
\caption{Bayesian Network of 238 counties at the same hour on two consecutive days in the training data (i.e., May 1st and May 2nd, 2018).}
\label{Fig:dag_vis}
\end{figure}

To evaluate our explanation, we visualize causal connections at different times in Figure~\ref{Fig:dag_vis}. Specifically, we show the Bayesian Network of 238 counties at the same hour on two consecutive days in the training data (i.e., May 1st and May 2nd, 2018). Interestingly, we can observe that two patterns in Figure~\ref{Fig:dag_vis} are almost identical at first glance. That could be the reason why DCRNN and GTS can perform well using the static structure. However, upon closer inspection, we find that these two are quite different to some extent if we zoom in, such as, in the upper right corner. Although the values have a tiny divergence, their volume is quite large. In two matrices of Figure~\ref{Fig:dag_vis}, the number of different cells is 28,509, and the corresponding percentage is $\frac{28509}{238 \times 238} \approx 0.5033$. We suppose that discovering those value-tiny but volume-big differences makes SGM outperform, to a large extent.

\subsection{Anomaly Detection}
After forecasting, we can have the hourly forecast of weather features at certain locations, denoted as $\mathcal{X}'$. Then, we use the encoder-decoder model in Figure~\ref{Fig:new_framework} to calculate the feature-wise generation probability using mean squared error (MSE) between $\mathcal{X}'$ and its generation $\bar{\mathcal{X}}'$. Thus, we can calculate the average of feature-wise generation probability as the condition of anomalies to identify if an anomaly weather pattern (e.g., a thunderstorm) happens in an hour in a particular location. In Table~\ref{tb:anomaly detection}, we use the Area Under the ROC Curve (i.e., AUC-ROC) as the metric, repeat the experiments four times, and report the performance of \name\ with baselines.

\begin{table*}[ht]
\caption{Anomaly Detection Performance (AUC-ROC)}
\centering
\scalebox{1}{
\begin{tabular}{ccccc}
\hline
         & NOAA-2017 $(\uparrow)$                    & NOAA-2018 $(\uparrow)$                    & NOAA-2019 $(\uparrow)$                    & NOAA-2020 $(\uparrow)$ \\ \hline
DeepSAD  & 0.5305 $\pm$ 0.0481                       & 0.5267 $\pm$ 0.0406                       & 0.5563 $\pm$ 0.0460                       & 0.6420 $\pm$ 0.0054           \\
DeepSVDD & 0.5201 $\pm$ 0.0045                       & 0.5603 $\pm$ 0.0111                       & $\textbf{0.6784}$ $\pm$ $\textbf{0.0112}$ & 0.5820 $\pm$ 0.0205           \\
DROCC    & 0.5319 $\pm$ 0.0661                       & 0.5103 $\pm$ 0.0147                       & 0.6236 $\pm$ 0.0992                       & 0.5630 $\pm$ 0.1082           \\ 
SGM   & $\textbf{0.5556}$ $\pm$ $\textbf{0.0010}$ & $\textbf{0.5685}$ $\pm$ $\textbf{0.0011}$ & 0.6298 $\pm$ 0.0184                       & $\textbf{0.6745}$ $\pm$ $\textbf{0.0185}$ \\ \hline
\end{tabular}}
\label{tb:anomaly detection}
\end{table*}

From Table~\ref{tb:anomaly detection}, we can observe that the detection module of \name\ achieves very competitive performance. An explanation is that, based on the anomalies distribution shown in Table~\ref{tb:label_distribution}, it can be observed that the anomalies are very rare events. Our generative manner could deal with the very rare scenario by learning the feature latent distributions instead of the (semi-)supervised learning manner.
For example, the maximum frequency of occurrences of thunderstorms is 770 (i.e., Jun 2017), which is collected from 238 counties over $30 \times 24 = 720$ hours, and the corresponding percentage is $\frac{770}{238 \times 30 \times 24} \approx 0.45\%$. 
Recall Remark~\ref{remark}, facing such rare events, we possibly find a single distribution to fit various anomaly patterns.

\subsection{Crop Classification}
In addition to the first two tasks, we also assess the quality of \name\ in the crop classification task. Table~\ref{tb:crop_classifiation} presents the results of baseline methods. We have the following observations: (1) All methods achieve good performance on some class, such as Open Water, Soybeans, Corn, Forest, etc, indicating the high quality of our benchmark. (2) These methods tend to perform worse in other classes, such as Sorghum, Other, Alfalfa. By investigation, we attribute this observation to the limited samples for these classes, comparing with the rich samples for the classes with good performance. (3) Our proposed method SGM outperforms baseline methods, demonstrating the effectiveness of the proposed method. 

\begin{table*}[ht]
\caption{Crop Classification}
\centering
\scalebox{1}{
\begin{tabular}{ccccccc cc}
\hline
 Baselines &  \multicolumn{2}{c}{SGM} &  \multicolumn{2}{c}{Swin} &  \multicolumn{2}{c}{DeepLabV3} \\ 
Classes                 & IoU ($\uparrow$)       & Acc ($\uparrow$)      & IoU ($\uparrow$)      & Acc ($\uparrow$)       & IoU ($\uparrow$)     & Acc ($\uparrow$)   \\ \hline
Natural Vegetation      & 39.23     & 46.86     &  45.66   &  71.80    & 47.31    & 64.28  \\ 
Forest                  & 42.44     & 61.07     &  34.47   &  41.63    & 46.50     & 77.10   \\ 
Corn                    & 53.30      & 63.56     &  52.00   &  62.53    & 52.30     & 72.81  \\ 
Soybeans                & 54.35     & 69.76     &  56.53   &  72.78    & 47.96    & 72.54  \\ 
Wetlands                & 40.17     & 59.55     &  42.15   &  69.57    & 35.42    & 43.62  \\ 
Developed/Barren        & 34.88     & 52.25     &  40.19   &  56.08    & 44.04    & 58.88  \\ 
Open Water              & 69.49     & 91.89     &  76.09   &  57.81   & 76.39    & 88.85  \\ 
Winter Wheat            & 55.54     & 75.96     &  48.21   &  86.41    & 47.75    & 54.32  \\ 
Alfalfa                 & 24.78     & 55.51     &  20.99   &  54.64    & 29.39    & 34.84  \\ 
Fallow/ Idle Cropland   & 38.32     & 61.75     &  37.14   &  23.23    & 17.55    & 19.45  \\ 
Cotton                  & 33.53     & 66.66     &  24.38   &  65.86     & 35.80     & 66.38  \\ 
Sorghum                 & 33.48     & 68.93     &  33.95   &  28.85     & 23.40     & 24.85  \\ 
Other                   & 28.27     & 42.81     &  28.72   &  45.56     & 27.14    & 41.58  \\ \hline
Average                 & \textbf{42.14}     & \textbf{62.81}     &  41.57   &      55.67     & 40.84    & 55.34  \\ 
\hline
\end{tabular}}
\label{tb:crop_classifiation}
\end{table*}

\section{Related Work}
\label{sec:related_work}


In recent years, there has been a surge in the development of benchmarking frameworks for weather and climate modeling methods.

For weather forecasting, WeatherBench~\cite{rasp2020weatherbench} utilizes datasets based on ERA5 archive and its updated version, WeatherBench 2~\cite{DBLP:journals/corr/abs-2308-15560}, provides evaluation frameworks with continually updated metrics and cutting-edge methods. As an extension of WeatherBench~\cite{rasp2020weatherbench}, WeatherBench Probability~\cite{DBLP:journals/corr/abs-2205-00865} supports probabilistic forecasting by adding established probabilistic verification metrics. For subseasonal weather forecasting, SubseasonalClimateUSA~\cite{DBLP:conf/nips/MouatadidOFOCWK23} proposes a benchmark dataset for a variety of subseasonal models. ClimART~\cite{DBLP:conf/nips/CachayRCBR21} presents a large dataset and challenging inference settings to benchmark the emulation of atmospheric radiative transfer of weather and climate models. ClimateBench~\cite{Watson-Parris2022-dz} provides a benchmark framework for machine learning models emulation of climate response to various emission scenarios. ClimateLearn~\cite{DBLP:conf/nips/NguyenJBSG23} offers an open-source and unified framework in dataset processing pipelines and model evaluation for various weather and climate modeling tasks.

In addition to the general climate benchmark mentioned above for weather forecasting, domain-specific climate settings also attract much research attention. For example, there are NADBenchmarks~\cite{DBLP:journals/corr/abs-2212-10735} for tasks related to natural disasters, as well as several datasets focusing on extreme weather events such as FloodNet~\cite{DBLP:journals/access/RahnemoonfarCSV21}, ExtremeWeather~\cite{DBLP:conf/nips/RacahBMKPP17}, EarthNet~\cite{DBLP:conf/cvpr/Requena-MesaBRR21}, DroughtED~\cite{minixhofer2021droughted}, and ClimateNet~\cite{Prabhat2021-ln}. Additionally, there are datasets targeting specific applications of climate science, such as cloud cluster classification~\cite{rasp2020combining}, storm classification~\cite{Haberlie2021-bi}, nowcasting~\cite{Franch2020-lo}, rain precipitation ~\cite{tong2020rainbench, DBLP:journals/corr/abs-2107-03432}, tropical cyclone intensity prediction~\cite{DBLP:journals/staeors/MaskeyRRGFKBCM20}, global air quality metrics estimation~\cite{Betancourt2021-rd}, and stream flow forecasting with flash flood damage estimation~\cite{DBLP:journals/corr/abs-2012-11154}.

Multi-modality naturally exists in climate modeling and many other domains~\cite{DBLP:journals/corr/abs-2410-17576, DBLP:conf/www/ZhouZZLH20, DBLP:conf/kdd/ZhengJLTH24,  DBLP:journals/corr/abs-2406-05375, DBLP:journals/corr/abs-2412-08174, DBLP:conf/sdm/ZhengZH23, DBLP:conf/kdd/ZhengXZH22, DBLP:conf/www/ZhengCYCH21, DBLP:conf/sdm/ZhengCH19, DBLP:journals/corr/abs-2409-09770}. Very recently, the trend of extending the modality of climate and weather benchmarks has emerged.
For example, in the natural language domain, ClimateX~\cite{DBLP:journals/corr/abs-2311-17107} proposes an expert-labeled dataset that comprises climate statements and their confidence levels. 

Motivated by the above analysis, we discerned that a multi-modality climate benchmark is interesting, and to the best of our knowledge, there is no related work proposed for this target. To this end, we propose our \name\ benchmark, which first aligns the ERA5 data for \textit{weather forecasting}, the NOAA data for \textit{thunderstorm alerts}, and HLS satellite image data for \textit{crop segmentation} based on the unified spatial-temporal granularity.

\section{Limitations and Future Directions}
\name\ represents an initial endeavor to establish a comprehensive multi-modality climate benchmark dataset aimed at fostering the development of next-generation AGI methodologies in climate science. While the current scope of \name\ encompasses multiple modalities, there remains significant potential for expansion.

A particularly promising direction is the integration of structured language representations that align with existing climate data\cite{li2025language}. For instance, as illustrated in Figure~\ref{Fig:geo_vis}, generating textual descriptions that accurately capture weather patterns and corresponding visual features presents a compelling research avenue with substantial scientific value. Moving forward, we plan to further enrich the multimodal capabilities of \name, with a particular emphasis on enhancing its language component~\cite{DBLP:journals/corr/abs-2410-12126}, recognizing its critical role in improving interpretability, accessibility, and downstream applications in climate research.

There are many different ways to model weather and climate changes, from analyzing and simulation of complex atmospheric physics~\cite{phillips1956general, lynch2008origins}, to data-driven techniques based on graphs and patterns~\cite{DBLP:conf/sigir/FuH21, keisler2022forecasting, DBLP:conf/kdd/FuFMTH22, lam2023learning, ben2024rise, DBLP:journals/corr/abs-2412-21151, li2024provably, li2024hypergraphs, DBLP:conf/iclr/FuHXFZSWMHL24, DBLP:journals/corr/abs-2410-13798}, spatiotemporal series and structures~\cite{karevan2020transductive, barrera2022rainfall, DBLP:conf/nips/BanZLQFKTH24, DBLP:conf/nips/TieuFZHH24, DBLP:conf/nips/Lin0FQT24, fang2025tsla}, foundation models~\cite{DBLP:conf/kdd/ZhengJLTH24, schmude2024prithvi, szwarcman2024prithvi}, knowledge-enhanced retrieval~\cite{DBLP:journals/corr/abs-2412-17336, juhasz2024responsible, schimanski2024climretrieve} and physics-informed networks~\cite{verma2024climode}. In this work, we mainly focus on the spatiotemporal series and foundation models based on satellite imagery, while other modeling techniques and modalities could be further incorporated in the future.


\section{Conclusion}
In conclusion, we provide a multi-modal climate benchmark named \name, integrating diverse datasets and assessing the quality of this benchmark by conducting experiments with various tasks. Our experimental results demonstrate the high quality of \name. Additionally, we propose SGM, a simple encoder-decoder-based generative model, which demonstrates competitive performance across various tasks.  These developments are crucial for improving climate modeling and prediction. By making this dataset publicly available, we aim to facilitate further research and innovation in multi-modal climate forecasting and anomaly detection, contributing significantly to the development of more robust and effective climate models.

\bibliographystyle{ACM-Reference-Format}
\bibliography{reference}

\appendix

\section{Refine Hidden Code by Location-wise Causality Discovery in SGM}
\label{sec:causality}

\begin{figure*}[h]
\includegraphics[width=0.94\textwidth]{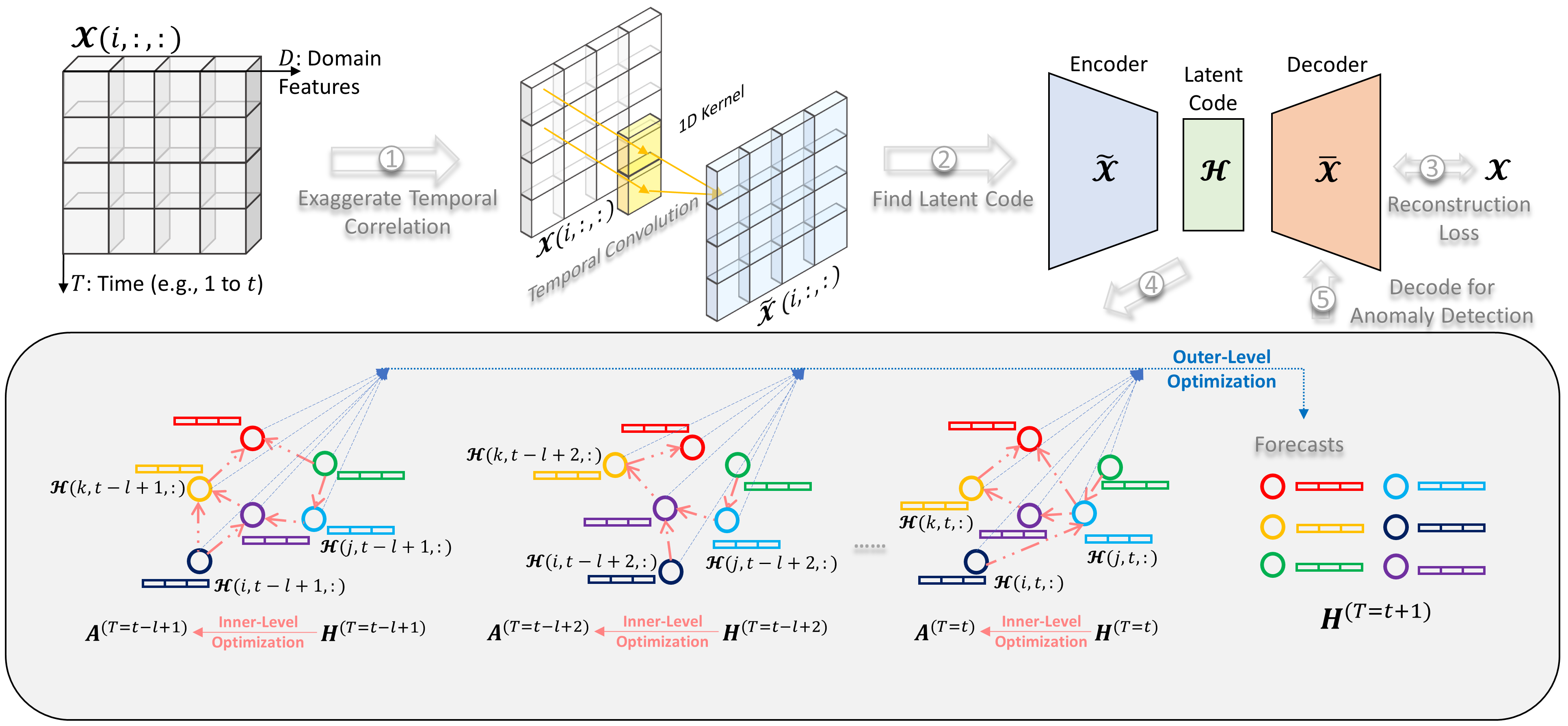}
\centering
\caption{Detailed Pipeline of Causality Discovery.}
\label{Fig:framework}
\end{figure*}

In this section, we introduce how SGM includes a causality discovery module by observing the historical tensor time series and utilizes it to guide tensor time series forecasting and anomaly detection. The overall procedures are shown in Figure~\ref{Fig:framework}.

\subsection{Overview}
The upper component of Figure~\ref{Fig:framework} represents the data preprocessing part (i.e., converting raw input $\mathcal{X}$ to latent representation $\mathcal{H}$) of SGM through a pre-trained autoencoder. The goal of this component is to leverage comprehensive causality to achieve seamless forecasting and anomaly detection.

The lower component of Figure~\ref{Fig:framework} shows how SGM discovers causality in the historical tensor time series (in the form of $\mathcal{H}$ other than $\mathcal{X}$) and generates future tensor time series. In brief, the optimization is bi-level. First, the inner optimization captures instantaneous effects among location variables at each timestamp. These causal structures are then stored in the form of a sequence of Bayesian Networks. Second, the outer optimization discovers the Neural Granger Causality among variables in a time window with the support of a sequence of Bayesian Networks.

\subsection{Inner Optimization for Identifying Instantaneous Causal Relations in Time Series}
Generally speaking, the inner optimization produces a sequence of Bayesian Networks for each observed timestamp. At time $t$, the instantaneous causality is discovered based on input features $\mathcal{H}(:,:,t) = \bm{H}^{(t)} \in \mathbb{R}^{N \times H}$, and is represented by a directed acyclic graph $\mathcal{G}^{(t)} = (\bm{A}^{(t)} \in \mathbb{R}^{N \times N}, \bm{H}^{(t)}\in \mathbb{R}^{N \times H})$. To be specific, $\bm{A}^{(t)}$ is a weighted adjacency matrix of the Bayesian Network at time $t$, and each cell represents the coefficient of causal effects between variables $u$ and $v \in \{1, \ldots, N\}$. The features (e.g., $\mathcal{H}(v,:,t)$) are transformed from the input raw features (e.g., $\mathcal{X}(v,:,t)$).

The reasoning for discovering the instantaneous causal effects in the form of the Bayesian Network originates from a widely adopted assumption of causal graph learning~\cite{DBLP:conf/nips/ZhengARX18, DBLP:conf/icml/YuCGY19, DBLP:journals/csur/GuoCLH020, DBLP:journals/corr/abs-2202-02195, gong2023rhino, DBLP:conf/www/ZhengCHC24, DBLP:journals/corr/abs-2410-10021}: there exists a ground-truth causal graph $\mathbf{S}^{(t)}$ that specifies instantaneous parents of variables to recover their value generating process. Therefore, in our inner optimization, the goal is to discover the causal structure $\mathbf{S}^{(t)}$ at each time $t$ by recovering the generation of input features $\bm{H}^{(t)}$. Specifically, given the observed $\bm{H}^{(t)}$, we aim to estimate a structure $\bm{A}^{(t)}$, through which a certain distribution $\bm{Z}^{(t)}$ could generate $\bm{H}^{(t)}$ for $t \in \{1, \ldots, T\}$. In this way, the instantaneous causal effects are discovered, and the corresponding structures are encoded in $\bm{A}^{(t)}$. The generation function is expressed as follows.
\begin{equation}
    \sum_{t} log \mathcal{P}(\bm{H}^{(t)}) = \sum_{t}log \int \mathcal{P}(\bm{H}^{(t)} | \bm{Z}^{(t)}) \mathcal{P}(\bm{Z}^{(t)}) d\bm{Z}^{(t)}
\label{eq: generative_model}
\end{equation}
where the generation likelihood $\mathcal{P}(\bm{H}^{(t)} | \bm{Z}^{(t)})$ also takes $\bm{A}^{(t)}$ as input. The complete formula is shown in Eq.~\ref{eq: DAG_Decoder}.

For Eq.~\ref{eq: generative_model}, on the one hand, it is hard to get the prior distribution $\mathcal{P}(\bm{Z}^{(t)})$, which is highly related to the distribution of ground-truth causal graph distribution $\mathcal{P}(\mathbf{S}^{(t)})$ at time $t$~\cite{DBLP:journals/corr/abs-2202-02195}. On the other hand, for the generation likelihood $\mathcal{P}(\bm{H}^{(t)} | \bm{Z}^{(t)})$, the actual posterior $\mathcal{P}(\bm{Z}^{(t)} | \bm{H}^{(t)})$ is also intractable. Thus, we resort to the variational autoencoder (VAE)~\cite{DBLP:journals/corr/KingmaW13}. In this way, the actual posterior $\mathcal{P}(\bm{Z}^{(t)} | \bm{H}^{(t)})$ can be replaced by the variational posterior $\mathcal{Q}(\bm{Z}^{(t)} | \bm{H}^{(t)})$, and the prior distribution $\mathcal{P}(\bm{Z}^{(t)})$ is approximated by a Gaussian distribution. Furthermore, the inside encoder and decoder modules should take the structure $\bm{A}^{(t)}$ as the input. This design can be realized by various off-the-shelf variational graph autoencoders such as VGAE~\cite{DBLP:journals/corr/KipfW16a}, etc. However, the inner optimization is coupled with the outer optimization, i.e., the instantaneous causality will be integrated with cross-time Granger causality to make inferences. The inner complex neural architectures and parameters may render the outer optimization module hard to train, especially when the outer module itself needs to be complex. Therefore, we extend the widely-adopted linear Structural Equation Model (SEM)~\cite{DBLP:conf/nips/ZhengARX18, DBLP:conf/icml/YuCGY19, DBLP:journals/corr/abs-2202-02195, gong2023rhino} to the time-respecting setting as follows.

For $\mathcal{Q}(\bm{Z}^{(t)} | \bm{H}^{(t)})$, the encoder equation is expressed as
\begin{equation}
    \bm{Z}^{(t)} = (\bm{I} - {\bm{A}^{(t)}}^{\top})f_{\theta_{enc}^{(t)}}(\bm{H}^{(t)})
\label{eq: DAG_Encoder}
\end{equation}
For $\mathcal{P}(\bm{H}^{(t)} | \bm{Z}^{(t)})$, the decoder equation is expressed as
\begin{equation}
    \bm{H}^{(t)} = f_{\theta_{dec}^{(t)}}((\bm{I} - {\bm{A}^{(t)}}^{\top})^{-1}\bm{Z}^{(t)})
\label{eq: DAG_Decoder}
\end{equation}
As analyzed above\footnote{The complete forms of $\mathcal{Q}(\bm{Z}^{(t)} | \bm{H}^{(t)})$ and $\mathcal{P}(\bm{H}^{(t)} | \bm{Z}^{(t)})$ are $\mathcal{Q}_{A^{(t)}}(\bm{Z}^{(t)} | \bm{H}^{(t)})$ and $\mathcal{P}_{A^{(t)}}(\bm{H}^{(t)} | \bm{Z}^{(t)})$, we omit the subscript ${A^{(t)}}$ for brevity.}, $f_{\theta_{enc}^{(t)}}$ and $f_{\theta_{dec}^{(t)}}$ do not need complicated neural architectures. Therefore, we can use two-layer MLPs for them.
Then, the objective function $\mathcal{L}_{DAG}^{(t)}$ for discovering the instantaneous causality at time $t$ is expressed as follows, which corresponds to the inner optimization.
\begin{equation}
\begin{split}
    \min\limits_{\theta^{(t)}_{enc},\theta^{(t)}_{dec}, \bm{A}^{(t)}} \mathcal{L}_{DAG}^{(t)} = ~D_{KL}(\mathcal{Q}(\bm{Z}^{(t)} | \bm{H}^{(t)}) \|& \mathcal{P}(\bm{Z}^{(t)}))\\ 
    - \mathbb{E}_{\mathcal{Q}(\bm{Z}^{(t)} | \bm{H}^{(t)})}[\text{log} \mathcal{P}(\bm{H}^{(t)} &| \bm{Z}^{(t)})] \\
     \text{s.t.}~ \sum_{t}\text{Tr}[(\bm{I} + \bm{A}^{(t)} \circ \bm{A}^{(t)})^{N}] - N =0, ~\text{for}~ t & \in \{1, \ldots, T\}
\end{split}
\label{eq: inner_opt}
\end{equation}
where the first term in $\mathcal{L}_{DAG}^{(t)}$ is the KL-divergence measuring the distance between the distribution of generated $\bm{Z}^{(t)}$ and the pre-defined Gaussian, and the second term is the reconstruction loss between the generated $\bm{Z}^{(t)}$ with the original input $\bm{H}^{(t)}$. Note that there is an important constraint, i.e.,  $\text{Tr}[(\bm{I} + \bm{A}^{(t)} \circ \bm{A}^{(t)})^{N}] - N =0$, on $\bm{A}^{(t)} \in \mathbb{R}^{N \times N}$. $\text{Tr} (\cdot)$ is the trace of a matrix, and $\circ$ denotes the Hadamard product. The meaning of the constraint is explained as follows.
The constraint in Eq.~\ref{eq: inner_opt}, i.e., $\text{Tr}[(\bm{I} + \bm{A}^{(t)} \circ \bm{A}^{(t)})^{N}] - N =0$ regularizes the acyclicity of $\bm{A}^{(t)}$ during the optimization process, i.e., the learned $\bm{A}^{(t)}$ should not have any possible closed-loops at any length.
\begin{lemma}
Let $\bm{A}^{(t)}$ be a weighted adjacency matrix (negative weights allowed). $\bm{A}^{(t)}$ has no $N$-length loops, if $\text{Tr}[(\bm{I} + \bm{A}^{(t)} \circ \bm{A}^{(t)})^{N}] - N =0$.
\label{eq:acyclicity}
\end{lemma}

The intuition is that there will be no $k$-length path from node $u$ to node $v$ on a binary adjacency matrix $\}(u,v) = 0$. Compared with original acyclicity constraints in ~\cite{DBLP:conf/icml/YuCGY19}, our Lemma~\ref{eq:acyclicity} gets rid of the $\lambda$ condition. Then we can denote $\alpha({A}^{(t)}) = \text{Tr}[(\bm{I} + \bm{A}^{(t)} \circ \bm{A}^{(t)})^{N}] - N$ and use Lagrangian optimization for Eq.~\ref{eq: inner_opt} as follows.
\begin{equation}
\begin{split}
    \min\limits_{\theta^{(t)}_{enc},\theta^{(t)}_{dec}, \bm{A}^{(t)}} \mathcal{L}_{DAG}^{(t)} =  ~D_{KL}(\mathcal{Q}(\bm{Z}^{(t)} | \bm{H}^{(t)}) \| \mathcal{P}(\bm{Z}^{(t)})) \\
    - \mathbb{E}_{\mathcal{Q}(\bm{Z}^{(t)} | \bm{H}^{(t)})}[\text{log} \mathcal{P}(\bm{H}^{(t)} | \bm{Z}^{(t)})] \\
    + \lambda ~\alpha({A}^{(t)}) + \frac{c}{2}|\alpha({A}^{(t)})|^{2},~ ~\text{for}~ t \in \{1, \ldots, T\}
\end{split}
\label{eq: complete_inner_opt}
\end{equation}
where $\lambda$ and $c$ are two hyperparameters, and larger $\lambda$ and $c$ enforce $\alpha(\bm{A}^{(t)})$ to be smaller.

\begin{theorem}
    If the ground-truth instantaneous causal graph $\mathbf{S}^{(t)}$ at time $t$ generates the features of variables following the normal distribution, then the inner optimization (i.e., Eq.~\ref{eq: inner_opt}) can identify $\mathbf{S}^{(t)}$ under the standard causal discovery assumptions~\cite{DBLP:journals/corr/abs-2202-02195}.
    \label{eq: discovery}
\end{theorem}

\subsection{Outer Optimization for Integrating Instantaneous Causality with Neural Granger}
Given the inner optimization, Bayesian Networks can be obtained at each timestamp $t$, which means that multiple instantaneous causalities are discovered. Thus, in the outer optimization, we integrate these evolving Bayesian Networks into Granger Causality discovery.
First, the classic Granger Causality~\cite{granger1969investigating} is discovered in the form of the variable-wise coefficients across different timestamps (i.e., a time window) through the autoregressive prediction process. The prediction based on the linear Granger Causality~\cite{granger1969investigating} is expressed as follows.
\begin{equation}
    \bm{H}^{(t)} = \sum_{l=1}^{L} \bm{W}^{(l)}\bm{H}^{(t-l)} + \bm{e}^{(t)}
\label{eq:graner_causality}
\end{equation}
where $\bm{H}^{(t)} \in \mathbb{R}^{N \times D}$ denotes the features of $N$ variables at time $t$, $\bm{e}^{(t)}$ is the noise, and $L$ is the pre-defined time lag indicating how many past timestamps can affect the values of $\bm{H}^{(t)}$. Weight matrix $\bm{W}^{(l)} \in \mathbb{R}^{N \times N}$ stores the cross-time coefficients captured by Granger Causality, i.e., matrix $\bm{W}^{(l)}$ aligns the variables at time $t-l$ with the variables at time $t$. To compute those weights, several linear methods are proposed, e.g., vector autoregressive model~\cite{DBLP:conf/kdd/ArnoldLA07}.

Facing non-linear causal relationships, neural Granger Causality discovery~\cite{DBLP:journals/pami/TankCFSF22} is recently proposed to explore the nonlinear Granger Causality effects. The general principle is to represent causal weights $\bm{W}$ by deep neural networks. To integrate instantaneous effects with neural Granger Causality discovery, our solution is expressed as follows.
\begin{equation}
    \hat{H}(i,:)^{(t)} =  f_{\Theta_{i}}[(\bm{A}^{(t-1)}, \bm{H}^{(t-1)}), \ldots, (\bm{A}^{(t-L)}, \bm{H}^{(t-L)})]
\end{equation}
where $L$ is the lag (or window size) in the Granger Causality, and $i$ is the index of the $i$-th variable. $f_{\Theta_{i}}$ is a neural computation unit with all parameters denoted as $\Theta_{i}$, whose input is an $L$-length time-ordered sequence of $(\bm{A}, \bm{H})$. And $f_{\Theta_{i}}$ is responsible for discovering the causality for variable $i$ at time $t$ from all variables that occurred in the past time lag $l$. The choice of neural unit $f_{\Theta_{i}}$ is flexible, such as MLP and LSTM~\cite{DBLP:journals/pami/TankCFSF22}. Different neural unit choices correspond to different causality interpretations. In our proposed SGM model, we use graph recurrent neural networks~\cite{DBLP:journals/tnn/WuPCLZY21}

We encode $f_{\Theta_{i}}$ into a sequence-to-sequence model~\cite{DBLP:conf/nips/SutskeverVL14}. That is, given a time window (or time lag), \name\ could forecast the corresponding features for the next time window.
Moreover, with $\bm{W}^{(l)}$ in Eq.~\ref{eq:graner_causality} and $f_{\Theta_{i}}$ in Eq.~\ref{eq:forecast}, we can observe that the classical linear Granger Causality $\bm{W}^{(l)}$ can be discovered for each time lag. In other words, each time lag has its own discovered coefficients, but $f_{\Theta_{i}}$ is shared by all time lags. This sharing manner is designed for scalability and is called Summary Causal Graph~\cite{DBLP:conf/iclr/MarcinkevicsV21, DBLP:conf/uai/AssaadDG22}. The underlying intuition is that the causal effects mainly depend on the near timestamps. Further, for the neural Granger Causality interpretation in $f_{\Theta_{i}}$, we follow the rule~\cite{DBLP:journals/pami/TankCFSF22} that if the $j$-th row of ($\bm{W}_{\bm{R} \ast \bm{A}^{(t)}}$, $\bm{W}_{\bm{C} \ast \bm{A}^{(t)}}$, and $\bm{W}_{\bm{U} \ast \bm{A}^{(t)}}$) are zeros, then variable $j$ is not the Granger-cause for variable $i$ in this time window.

In the outer optimization, to evaluate the prediction, we use the mean absolute error (MAE) loss on the prediction and the ground truth, which is effective and widely applied to time-series forecasting tasks~\cite{DBLP:conf/iclr/LiYS018, DBLP:conf/iclr/Shang0B21}.
\begin{equation}
   \min_{\Theta_{i}, \bm{A}^{(t-1)}, \ldots, \bm{A}^{(t-l)} } \mathcal{L}_{pred} = \sum_{i} \sum_{t} | H(i,:)^{(t)} - \hat{H}(i,:)^{(t)}|
\end{equation}
where $\Theta_{i}, \bm{A}^{(t-1)}, \ldots, \bm{A}^{(t-l)}$ are all the parameters for the prediction $\hat{H}(i,:)^{(t)}$ of variable $i$ at time $t$. The composition and update rules are expressed below.

\textbf{For updating $f_{\Theta_i}$}, we employ the recurrent neural structure to fit the input sequence. Moreover, the sequential inputs also contain the structured data $\bm{A}$. Therefore, we use the graph recurrent neural architecture~\cite{DBLP:conf/iclr/LiYS018} because it is designed for directed graphs, whose core is a gated recurrent unit~\cite{DBLP:journals/corr/ChungGCB14}.
\begin{equation}
\begin{split}
    & \bm{R}^{(t)} = \text{sigmoid} (\bm{W}_{\bm{R} \ast \bm{A}^{(t)}} [\bm{H}^{(t)} \oplus \bm{S}^{(t-1)}] + \bm{b}_{R})\\
    & \bm{C}^{(t)} = \text{tanh} (\bm{W}_{\bm{C} \ast \bm{A}^{(t)}} [\bm{H}^{(t)} \oplus (\bm{R}^{(t)} \odot \bm{S}^{(t-1)})] + \bm{b}_{C})\\
    & \bm{U}^{(t)} = \text{sigmoid} (\bm{W}_{\bm{U} \ast \bm{A}^{(t)}} [\bm{H}^{(t)} \oplus \bm{S}^{(t-1)}] + \bm{b}_{U})\\
    & \bm{S}^{(t)} = \bm{U}^{(t)} \odot \bm{S}^{(t-1)} + (\bm{I} - \bm{U}^{(t)}) \odot \bm{C}^{(t)}
\end{split}
\label{eq: gates}
\end{equation}
where $\bm{R}^{(t)}$, $\bm{C}^{(t)}$, and $\bm{U}^{(t)}$ are three parameterized gates, with corresponding weights $\bm{W}$ and bias $\bm{b}$. $\bm{H}^{(t)}$ is the input, and $\bm{S}^{(t)}$ is the hidden state. Gates $\bm{R}^{(t)}$, $\bm{C}^{(t)}$, and $\bm{U}^{(t)}$ share the similar structures. For example, in $\bm{R}^{(t)}$, the graph convolution operation for computing the weight $\bm{W}_{\bm{R} \ast \bm{A}^{(t)}}$ is defined as follows, and the same computation applies to gates $\bm{U}^{(t)}$ and $\bm{C}^{(t)}$.
\begin{equation}
    \bm{W}_{\bm{R} \ast \bm{A}^{(t)}} = \sum_{k=0}^{K} ~\theta_{k,1}^{R}({\bm{D}^{(t)}_{out}}^{-1}\bm{A}^{(t)})^{k} +\theta_{k,2}^{R}({\bm{D}^{(t)}_{in}}^{-1}{\bm{A}^{(t)}}^{\top})^{k}
\end{equation}
where $\theta_{k,1}^{R}$, $\theta_{k,2}^{R}$ are learnable weight parameters; scalar $k$ is the order for the stochastic diffusion operation (i.e., similar to steps of random walks); ${\bm{D}^{(t)}_{out}}^{-1}\bm{A}^{(t)}$ and ${\bm{D}^{(t)}_{in}}^{-1}{\bm{A}^{(t)}}^{\top}$ serve as the transition matrices with the in-degree matrix $\bm{D}^{(t)}_{in}$ and the out-degree matrix $\bm{D}^{(t)}_{out}$; $-1$ and $\top$ are inverse and transpose operations.

\textbf{For updating each of $\{\bm{A}^{(t-1)}, \ldots, \bm{A}^{(t-l)}\}$}, we take $\bm{A}^{(t-l)}$ as an example to illustrate. The optimal $\bm{A}^{(t-l)}$ stays in the space of $\{0,1\}^{N \times N}$. To be specific, each edge $A^{(t-l)}(i,j)$ can be parameterized as $\theta_{i,j}^{(t-l)}$ following the Bernoulli distribution. However, $N^{2}l$ is hard to scale, and the discrete variables are not differentiable. Therefore, we adopt the Gumbel reparameterization from~\cite{DBLP:conf/iclr/JangGP17, DBLP:conf/iclr/MaddisonMT17}. It provides a continuous approximation for the discrete distribution, which has been widely used in the graph structure learning~\cite{DBLP:conf/icml/KipfFWWZ18, DBLP:conf/iclr/Shang0B21}. The general reparameterization form can be written as $A^{(t-l)}(i,j) = softmax (FC((H(i,:)^{(t-l)}||H(j,:)^{(t-l)}) + g)/\xi)$, where $FC$ is a feedforward neural network, $g$ is a scalar drawn from a Gumbel$(0,1)$ distribution, and $\xi$ is a scaling hyperparameter. Different from~\cite{DBLP:conf/icml/KipfFWWZ18, DBLP:conf/iclr/Shang0B21}, in our setting, the initial structure input is constrained by the causality discovery, which originates from the inner optimization step. Hence, the structure learning in the outer optimization takes the adjacency matrix from the inner optimization as the initial input, which is
\begin{equation}
    A^{(t-l)}_{outer}(i,j) = softmax (A^{(t-l)}_{inner}(i,j) + \bm{g})/\xi)
\end{equation}
where $A^{(t)}_{inner}(i,j)$ is the structure learned by our inner optimization through Eq.~\ref{eq: inner_opt}, $A^{(t)}_{outer}(i,j)$ is the updated structure, and $\bm{g}$ is a vector of i.i.d samples drawn from a Gumbel$(0,1)$ distribution. In outer optimization, Eq.~\ref{eq: outer_opt} fine-tunes the evolving Bayesian Networks to make the intra-time causality fit the cross-time causality well. Note that, the outer optimization w.r.t. $\bm{A}^{(t)}$ may break the acyclicity, and another round of inner optimization may be necessary.

\subsection{Model-agnostic Autoencoder}
As shown in Figure~\ref{Fig:framework}, the autoencoder can be pre-trained with reconstruction loss (e.g., MSE) ahead of the inner and outer optimization, to obtain $\mathcal{H}$ for the feature latent distribution representation. By utilizing all input $\mathcal{H}$, the inner optimization learns the sequential Bayesian Networks, and the outer optimization aligns Bayesian Networks with the neural Granger Causality to produce all the forecast $\mathcal{H}'$. The inner and outer optimization can be trained interchangeably.

\subsection{Theoretical Analysis}
\label{sec:theoretical_analysis}

\subsubsection{Proof of Lemma~\ref{eq:acyclicity}}
Following~\cite{DBLP:conf/icml/YuCGY19}, at each time $t$, we can extend
$(\bm{I} + \bm{A}^{(t)} \circ \bm{A}^{(t)})^{N}$ by binomial expansion as follows.

\begin{equation}
    (\bm{I} + \bm{A}^{(t)} \circ \bm{A}^{(t)})^{N} = \bm{I} + \sum_{k=1}^{N} \begin{pmatrix} N \\ k \end{pmatrix} (\bm{A}^{(t)})^{k}
\end{equation}

Since 
\begin{equation}
    \bm{I} \in \mathbb{R}^{N \times N}
\end{equation}
then
\begin{equation}
    \text{Tr}(\bm{I}) = N
\end{equation}

Thus, if
\begin{equation}
    (\bm{I} + \bm{A}^{(t)} \circ \bm{A}^{(t)})^{N} - N = 0
\end{equation}
then
\begin{equation}
    (\bm{A}^{(t)})^{k} = 0, \text{~for any~} k
\end{equation}

Therefore, $\bm{A}^{(t)}$ is acyclic, i.e., no closed-loop exists in $\bm{A}^{(t)}$ at any possible length. Overall, the general idea of Lemma~\ref{eq:acyclicity} is to ensure that the diagonal entries of the powered adjacency matrix have no $1$s. There are also other forms for acyclicity constraints obeying the same idea but in different expressions, like exponential power form in~\cite{DBLP:conf/nips/ZhengARX18}.

\subsubsection{Sketch Proof of Theorem~\ref{eq: discovery}}
According to Theorem 1 from~\cite{DBLP:journals/corr/abs-2202-02195}, the ELBO form as our Eq.~\ref{eq: inner_opt} could identity the ground-truth causal structure $\mathbf{S}^{(t)}$ at each time $t$. The difference between our ELBO and the ELBO in~\cite{DBLP:journals/corr/abs-2202-02195} is entries in the KL-divergence. Specifically, in~\cite{DBLP:journals/corr/abs-2202-02195}, the prior and variational posterior distributions are on the graph level. Usually, the prior distribution of graph structures is not easy to obtain (e.g., the non-IID and heterophyllous properties). Then, we transfer the graph structure distribution to the feature distribution that the Gaussian distribution can model. That's why our prior and variational posterior distributions in the KL-divergence are on the feature (generated by the graph) level.

\subsection{Implementation}
\label{sec:implementation}

\subsubsection{Hyperparameter Search}
In Eq.~\ref{eq: complete_inner_opt}, instead of fixing the hyperparameter $\lambda$ and $c$ during the optimization. Increasing the values of hyperparameter $\lambda$ and $c$ can reduce the possibility that learned structures break the acyclicity~\cite{DBLP:conf/icml/YuCGY19}, such that one iterative way to increase hyperparameters $\lambda$ and $c$ during the optimization can be expressed as follows.
\begin{equation}
    \lambda_{i+1} \leftarrow \lambda_{i} + c_{i} \alpha(\bm{A}^{(t)}_{i})
\end{equation}
and
\begin{equation}  
    c_{i+1} = \begin{cases}
    \eta c_{i} &\text {if $|\alpha(\bm{A}^{(t)}_{i})| > \gamma |\alpha(\bm{A}^{(t)}_{i-1})|$}\\
    c_{i} &\text {otherwise}
    \end{cases}
\end{equation}
where $\eta > 1$ and $0 < \gamma < 1$ are two hyperparameters, the condition $|\alpha(\bm{A}^{(t)}_{i})| > \gamma |\alpha(\bm{A}^{(t)}_{i-1})|$ means that the current acyclicity $\alpha(\bm{A}^{(t)}_{i})$ at the $i$-th iteration is not ideal, because it is not decreased below the $\gamma$ portion of $\alpha(\bm{A}^{(t)}_{i-1})$ from the last iteration $i-1$.

\subsubsection{Reproducibility}
For forecasting and anomaly detection, we have four cross-validation groups. For example, focusing on an interesting time interval each year (e.g., from May to August is the season for frequent thunderstorms), we set group \#1 with [2018, 2019, 2020] as training, [2021] as validation, and [2017] as testing. Thus, we have 8856 hours, 45 weather features, and 238 counties in the training set. The rest three groups are \{[2019, 2020, 2021], [2017], [2018]\}, \{[2020, 2021, 2017], [2018], [2019]\}, and \{[2021, 2017, 2018], [2019], [2020]\}, respectively. Therefore, SGM and baselines are required to forecast the testing set and detect the anomaly patterns in the testing set.

The persistence forecasting can be expressed as 
\begin{equation}
    \bm{X}^{(t)}_{SGM++} = \alpha \bm{X}^{(t)}_{SGM} + (1-\alpha) \bm{X}^{(t-\tau)} ~\text {~s.t.~} \bm{X}^{(t)}_{SGM} = \text{SGM}(\bm{X}^{(t-\tau)})
\end{equation}
where $\tau$ is the time window, for example, in the experiments, $\tau$ = 24h. $\text{SGM}(\bm{X}^{(t-\tau)})$ denotes the forecast of applying SGM on the input $\bm{X}^{(t-\tau)}$.

The synthetic data is publicly available~\footnote{\url{https://github.com/i6092467/GVAR}}. According to the corporate policy, our contributed data and the code of SGM will be released after the paper is published. The experiments are programmed based on Python and Pytorch on a Windows machine with 64GB RAM and a 16GB RTX 5000 GPU.

\onecolumn
\section{Feature Description of the Time Series Data in \name}
\label{weather_feature_description}

\begin{longtable}{p{2.5cm}p{2cm}p{6cm}p{2.5cm}}
\caption{Feature Descriptions with Instance Values Sampled from Jefferson, Alabama U.S. on 9:00-10:00, 01/05/2017, UTC}\\
    \toprule
    Feature    & Unit & Description & Value\\
    \midrule
    100-meter wind towards east  &m s$^{-1}$ &This parameter is the eastward component of the 100 m wind. It is the horizontal speed of air moving towards the east, at a height of 100 meters above the surface of the Earth, in meters per second. Care should be taken when comparing model parameters with observations, because observations are often local to a particular point in space and time, rather than representing averages over a model grid box. This parameter can be combined with the northward component to give the speed and direction of the horizontal 100 m wind. & -3.192476\\
    \midrule
    100-meter wind towards north  &m s$^{-1}$ &This parameter is the northward component of the 100 m wind. It is the horizontal speed of air moving towards the north, at a height of 100 meters above the surface of the Earth, in meters per second. Care should be taken when comparing model parameters with observations, because observations are often local to a particular point in space and time, rather than representing averages over a model grid box. This parameter can be combined with the eastward component to give the speed and direction of the horizontal 100 m wind. & -1.892055\\
    \midrule
    10-meter wind gust (maximum)  &m s$^{-1}$ &Maximum 3-second wind at 10 m height as defined by WMO. Parametrization represents turbulence only before 01102008; thereafter effects of convection are included. The 3 s gust is computed every time step, and the maximum is kept since the last postprocessing. & 3.620435\\
    \midrule
    10-meter wind gust (instantaneous)  &m s$^{-1}$ &This parameter is the maximum wind gust at the specified time, at a height of ten meters above the surface of the Earth. The WMO defines a wind gust as the maximum of the wind averaged over 3-second intervals. This duration is shorter than a model time step, and so the ECMWF Integrated Forecasting System (IFS) deduces the magnitude of a gust within each time step from the time-step-averaged surface stress, surface friction, wind shear, and stability. Care should be taken when comparing model parameters with observations, because observations are often local to a particular point in space and time, rather than representing averages over a model grid box. & 3.178461\\
    \midrule
    10-meter wind towards east   &m s$^{-1}$ &	This parameter is the eastward component of the 10m wind. It is the horizontal speed of air moving towards the east, at a height of ten meters above the surface of the Earth, in meters per second. Care should be taken when comparing this parameter with observations because wind observations vary on small space and time scales and are affected by the local terrain, vegetation, and buildings that are represented only on average in the ECMWF Integrated Forecasting System (IFS). This parameter can be combined with the V component of 10m wind to give the speed and direction of the horizontal 10m wind. & -1.094084\\
    \midrule
    10-meter wind towards north   &m s$^{-1}$ &	This parameter is the northward component of the 10m wind. It is the horizontal speed of air moving towards the north, at a height of ten metres above the surface of the Earth, in metres per second. Care should be taken when comparing this parameter with observations, because wind observations vary on small space and time scales and are affected by the local terrain, vegetation and buildings that are represented only on average in the ECMWF Integrated Forecasting System (IFS). This parameter can be combined with the U component of 10m wind to give the speed and direction of the horizontal 10m wind. & -1.119224\\
    \midrule
    Atmospheric water content   & kg m$^{-2}$ & This parameter is the sum of water vapor, liquid water, cloud ice, rain, and snow in a column extending from the surface of the Earth to the top of the atmosphere. In old versions of the ECMWF model (IFS), rain and snow were not accounted for. & 9.287734\\
    \midrule
    Atmospheric water vapor content   & kg m$^{-2}$ & This parameter is the total amount of water vapor in a column extending from the surface of the Earth to the top of the atmosphere. This parameter represents the area averaged value for a grid box. & 9.287452\\
    \midrule
    Dewpoint    & K & This parameter is the temperature to which the air, at 2 meters above the surface of the Earth, would have to be cooled for saturation to occur. It is a measure of the humidity of the air. Combined with temperature and pressure, it can be used to calculate relative humidity. 2m dew point temperature is calculated by interpolating between the lowest model level and the Earth's surface, taking account of the atmospheric conditions. This parameter has units of kelvin (K). Temperature measured in kelvin can be converted to degrees Celsius (°C) by subtracting 273.15. & 269.059570\\
    \midrule
    High cloud cover    & Dimensionless & The proportion of a grid box covered by cloud occurring in the high levels of the troposphere. High cloud is a single-level field calculated from cloud occurring on model levels with a pressure less than 0.45 times the surface pressure. So, if the surface pressure is 1000 hPa (hectopascal), high cloud would be calculated using levels with a pressure of less than 450 hPa (approximately 6km and above (assuming a "standard atmosphere")). The high cloud cover parameter is calculated from the cloud for the appropriate model levels described above. Assumptions are made about the degree of overlap/randomness between clouds in different model levels. Cloud fractions vary from 0 to 1. & 0.224129\\
    \midrule
    Low cloud cover    & Dimensionless & This parameter is the proportion of a grid box covered by cloud occurring in the lower levels of the troposphere. Low cloud is a single level field calculated from cloud occurring on model levels with a pressure greater than 0.8 times the surface pressure. So, if the surface pressure is 1000 hPa (hectopascal), low cloud would be calculated using levels with a pressure greater than 800 hPa (below approximately 2km (assuming a "standard atmosphere")). Assumptions are made about the degree of overlap/randomness between clouds in different model levels. This parameter has values from 0 to 1. & 0.000000\\
    \midrule
    Gravitational potential energy    & m$^{2}$ s$^{-2}$ & This parameter is the gravitational potential energy of a unit mass, at a particular location at the surface of the Earth, relative to mean sea level. It is also the amount of work that would have to be done, against the force of gravity, to lift a unit mass to that location from mean sea level. The (surface) geopotential height (orography) can be calculated by dividing the (surface) geopotential by the Earth's gravitational acceleration, g (=9.80665 m s-2 ). This parameter does not vary in time. &NaN\\
    \midrule
    Medium cloud cover     & Dimensionless & This parameter is the proportion of a grid box covered by cloud occurring in the middle levels of the troposphere. Medium cloud is a single level field calculated from cloud occurring on model levels with a pressure between 0.45 and 0.8 times the surface pressure. So, if the surface pressure is 1000 hPa (hectopascal), medium cloud would be calculated using levels with a pressure of less than or equal to 800 hPa and greater than or equal to 450 hPa (between approximately 2km and 6km (assuming a "standard atmosphere")). The medium cloud parameter is calculated from cloud cover for the appropriate model levels as described above. Assumptions are made about the degree of overlap/randomness between clouds in different model levels. Cloud fractions vary from 0 to 1. &0.000000\\
    \midrule
    Maximum temperature      & k & This parameter is the highest temperature of air at 2m above the surface of land, sea or inland water since the parameter was last archived in a particular forecast. 2m temperature is calculated by interpolating between the lowest model level and the Earth's surface, taking account of the atmospheric conditions. This parameter has units of kelvin (K). Temperature measured in kelvin can be converted to degrees Celsius (°C) by subtracting 273.15. &273.357666\\
    \midrule
    Maximum precipitation rate      & kg m$^{-2}$ s$^{-1}$ & The total precipitation is calculated from the combined large-scale and convective rainfall and snowfall rates every time step and the maximum is kept since the last postprocessing. &0.000000\\
    \midrule
    Mean sea level pressure      & Pa & This parameter is the pressure (force per unit area) of the atmosphere at the surface of the Earth, adjusted to the height of mean sea level. It is a measure of the weight that all the air in a column vertically above a point on the Earth's surface would have, if the point were located at mean sea level. It is calculated over all surfaces - land, sea and inland water. Maps of mean sea level pressure are used to identify the locations of low and high pressure weather systems, often referred to as cyclones and anticyclones. Contours of mean sea level pressure also indicate the strength of the wind. Tightly packed contours show stronger winds. The units of this parameter are pascals (Pa). Mean sea level pressure is often measured in hPa and sometimes is presented in the old units of millibars, mb (1 hPa = 1 mb = 100 Pa). &101550.976562\\
    \midrule
    Minimum temperature       & k & This parameter is the lowest temperature of air at 2m above the surface of land, sea or inland waters since the parameter was last archived in a particular forecast. 2m temperature is calculated by interpolating between the lowest model level and the Earth's surface, taking account of the atmospheric conditions. See further information. This parameter has units of kelvin (K). Temperature measured in kelvin can be converted to degrees Celsius (°C) by subtracting 273.15. &273.357666\\
    \midrule
    Minimum precipitation rate      & kg m$^{-2}$ s$^{-1}$ & The total precipitation is calculated from the combined large-scale and convective rainfall and snowfall rates every time step and the minimum is kept since the last postprocessing. &0.000000\\
    \midrule
    Precipitation type & Dimensionless & This parameter describes the type of precipitation at the surface, at the specified time. A precipitation type is assigned wherever there is a non-zero value of precipitation. The ECMWF Integrated Forecasting System (IFS) has only two predicted precipitation variables: rain and snow. Precipitation type is derived from these two predicted variables in combination with atmospheric conditions, such as temperature. Values of precipitation type defined in the IFS: 0: No precipitation, 1: Rain, 3: Freezing rain (i.e. supercooled raindrops which freeze on contact with the ground and other surfaces), 5: Snow, 6: Wet snow (i.e. snow particles which are starting to melt); 7: Mixture of rain and snow, 8: Ice pellets. These precipitation types are consistent with WMO Code Table 4.201. Other types in this WMO table are not defined in the IFS. &0.000000\\
    \midrule
    Rain water content of atmosphere & kg m$^{-2}$ &This parameter is the total amount of water in droplets of raindrop size (which can fall to the surface as precipitation) in a column extending from the surface of the Earth to the top of the atmosphere. This parameter represents the area averaged value for a grid box. Clouds contain a continuum of different sized water droplets and ice particles. The ECMWF Integrated Forecasting System (IFS) cloud scheme simplifies this to represent a number of discrete cloud droplets/particles including: cloud water droplets, raindrops, ice crystals and snow (aggregated ice crystals). Droplet formation, conversion and aggregation processes are also highly simplified in the IFS. 0.000000 &0.000000\\
    \midrule
    Snow density & kg m$^{-3}$ &This parameter is the mass of snow per cubic metre in the snow layer. The ECMWF Integrated Forecasting System (IFS) represents snow as a single additional layer over the uppermost soil level. The snow may cover all or part of the grid box. This parameter is defined over the whole globe, even where there is no snow. Regions without snow can be masked out by only considering grid points where the snow depth (m of water equivalent) is greater than 0.0. &99.999985\\
    \midrule
    Snow depth & m of water equivalent &This parameter is the amount of snow from the snow-covered area of a grid box. Its units are metres of water equivalent, so it is the depth the water would have if the snow melted and was spread evenly over the whole grid box. The ECMWF Integrated Forecasting System (IFS) represents snow as a single additional layer over the uppermost soil level. The snow may cover all or part of the grid box. &0.000000\\
    \midrule
    Snowfall  & m of water equivalent &This parameter is the accumulated snow that falls to the Earth's surface. It is the sum of large-scale snowfall and convective snowfall. Large-scale snowfall is generated by the cloud scheme in the ECMWF Integrated Forecasting System (IFS). The cloud scheme represents the formation and dissipation of clouds and large-scale precipitation due to changes in atmospheric quantities (such as pressure, temperature and moisture) predicted directly at spatial scales of the grid box or larger. Convective snowfall is generated by the convection scheme in the IFS, which represents convection at spatial scales smaller than the grid box. In the IFS, precipitation is comprised of rain and snow. This parameter is accumulated over a particular time period which depends on the data extracted. For the reanalysis, the accumulation period is over the 1 hour ending at the validity date and time. For the ensemble members, ensemble mean and ensemble spread, the accumulation period is over the 3 hours ending at the validity date and time. The units of this parameter are depth in metres of water equivalent. It is the depth the water would have if it were spread evenly over the grid box. Care should be taken when comparing model parameters with observations, because observations are often local to a particular point in space and time, rather than representing averages over a model grid box. &0.000000\\
    \midrule
    Soil temperature (0 to 7 cm) &K & This parameter is the temperature of the soil at level 1 (in the middle of layer 1). The ECMWF Integrated Forecasting System (IFS) has a four-layer representation of soil, where the surface is at 0cm: Layer 1: 0 - 7cm, Layer 2: 7 - 28cm, Layer 3: 28 - 100cm, Layer 4: 100 - 289cm. Soil temperature is set at the middle of each layer, and heat transfer is calculated at the interfaces between them. It is assumed that there is no heat transfer out of the bottom of the lowest layer. Soil temperature is defined over the whole globe, even over ocean. Regions with a water surface can be masked out by only considering grid points where the land-sea mask has a value greater than 0.5. This parameter has units of kelvin (K). Temperature measured in kelvin can be converted to degrees Celsius (°C) by subtracting 273.15. &276.865784\\
    \midrule
    Soil temperature (7 to 28 cm)  &K & This parameter is the temperature of the soil at level 2 (in the middle of layer 2). The ECMWF Integrated Forecasting System (IFS) has a four-layer representation of soil, where the surface is at 0cm: Layer 1: 0 - 7cm, Layer 2: 7 - 28cm, Layer 3: 28 - 100cm, Layer 4: 100 - 289cm. Soil temperature is set at the middle of each layer, and heat transfer is calculated at the interfaces between them. It is assumed that there is no heat transfer out of the bottom of the lowest layer. Soil temperature is defined over the whole globe, even over ocean. Regions with a water surface can be masked out by only considering grid points where the land-sea mask has a value greater than 0.5. This parameter has units of kelvin (K). Temperature measured in kelvin can be converted to degrees Celsius (°C) by subtracting 273.15. &282.708038\\
    \midrule
    Soil temperature (28 to 100 cm)  &K & This parameter is the temperature of the soil at level 3 (in the middle of layer 3). The ECMWF Integrated Forecasting System (IFS) has a four-layer representation of soil, where the surface is at 0cm: Layer 1: 0 - 7cm, Layer 2: 7 - 28cm, Layer 3: 28 - 100cm, Layer 4: 100 - 289cm. Soil temperature is set at the middle of each layer, and heat transfer is calculated at the interfaces between them. It is assumed that there is no heat transfer out of the bottom of the lowest layer. Soil temperature is defined over the whole globe, even over ocean. Regions with a water surface can be masked out by only considering grid points where the land-sea mask has a value greater than 0.5. This parameter has units of kelvin (K). Temperature measured in kelvin can be converted to degrees Celsius (°C) by subtracting 273.15. &286.920227\\
    \midrule
    Soil temperature (100 to 289 cm)  &K & This parameter is the temperature of the soil at level 4 (in the middle of layer 4). The ECMWF Integrated Forecasting System (IFS) has a four-layer representation of soil, where the surface is at 0cm: Layer 1: 0 - 7cm, Layer 2: 7 - 28cm, Layer 3: 28 - 100cm, Layer 4: 100 - 289cm. Soil temperature is set at the middle of each layer, and heat transfer is calculated at the interfaces between them. It is assumed that there is no heat transfer out of the bottom of the lowest layer. Soil temperature is defined over the whole globe, even over ocean. Regions with a water surface can be masked out by only considering grid points where the land-sea mask has a value greater than 0.5. This parameter has units of kelvin (K). Temperature measured in kelvin can be converted to degrees Celsius (°C) by subtracting 273.15. &290.265320\\
    \midrule
    Snow water content of atmosphere & k m$^{-2}$ & This parameter is the total amount of water in the form of snow (aggregated ice crystals which can fall to the surface as precipitation) in a column extending from the surface of the Earth to the top of the atmosphere. This parameter represents the area averaged value for a grid box. Clouds contain a continuum of different sized water droplets and ice particles. The ECMWF Integrated Forecasting System (IFS) cloud scheme simplifies this to represent a number of discrete cloud droplets/particles including: cloud water droplets, raindrops, ice crystals and snow (aggregated ice crystals). Droplet formation, conversion and aggregation processes are also highly simplified in the IFS. &0.000069\\
    \midrule
    Soil water (0 to 7 cm)  & m$^{3}m^{-3}$ & This parameter is the volume of water in soil layer 1 (0 - 7cm, the surface is at 0cm). The ECMWF Integrated Forecasting System (IFS) has a four-layer representation of soil: Layer 1: 0 - 7cm, Layer 2: 7 - 28cm, Layer 3: 28 - 100cm, Layer 4: 100 - 289cm. Soil water is defined over the whole globe, even over ocean. Regions with a water surface can be masked out by only considering grid points where the land-sea mask has a value greater than 0.5. The volumetric soil water is associated with the soil texture (or classification), soil depth, and the underlying groundwater level. &0.439442\\
    \midrule
    Soil water (7 to 28 cm)  & m$^{3}m^{-3}$ & This parameter is the volume of water in soil layer 2 (7 - 28cm, the surface is at 0cm). The ECMWF Integrated Forecasting System (IFS) has a four-layer representation of soil: Layer 1: 0 - 7cm, Layer 2: 7 - 28cm, Layer 3: 28 - 100cm, Layer 4: 100 - 289cm. Soil water is defined over the whole globe, even over ocean. Regions with a water surface can be masked out by only considering grid points where the land-sea mask has a value greater than 0.5. The volumetric soil water is associated with the soil texture (or classification), soil depth, and the underlying groundwater level. &0.447512\\
    \midrule
    Soil water (28 to 100 cm)   & m$^{3}m^{-3}$ & This parameter is the volume of water in soil layer 3 (28 - 100cm, the surface is at 0cm). The ECMWF Integrated Forecasting System (IFS) has a four-layer representation of soil: Layer 1: 0 - 7cm, Layer 2: 7 - 28cm, Layer 3: 28 - 100cm, Layer 4: 100 - 289cm. Soil water is defined over the whole globe, even over ocean. Regions with a water surface can be masked out by only considering grid points where the land-sea mask has a value greater than 0.5. The volumetric soil water is associated with the soil texture (or classification), soil depth, and the underlying groundwater level. &0.387898\\
    \midrule
    Soil water (100 to 289 cm)   & m$^{3}m^{-3}$ & This parameter is the volume of water in soil layer 4 (100 - 289cm, the surface is at 0cm). The ECMWF Integrated Forecasting System (IFS) has a four-layer representation of soil: Layer 1: 0 - 7cm, Layer 2: 7 - 28cm, Layer 3: 28 - 100cm, Layer 4: 100 - 289cm. Soil water is defined over the whole globe, even over ocean. Regions with a water surface can be masked out by only considering grid points where the land-sea mask has a value greater than 0.5. The volumetric soil water is associated with the soil texture (or classification), soil depth, and the underlying groundwater level. &0.380035\\
    \midrule
    Solar radiation    & Jm$^{-2}$ & This parameter is the amount of solar radiation (also known as shortwave radiation) that reaches a horizontal plane at the surface of the Earth. This parameter comprises both direct and diffuse solar radiation. 
    
    Radiation from the Sun (solar, or shortwave, radiation) is partly reflected back to space by clouds and particles in the atmosphere (aerosols) and some of it is absorbed. The rest is incident on the Earth's surface (represented by this parameter).
    
    To a reasonably good approximation, this parameter is the model equivalent of what would be measured by a pyranometer (an instrument used for measuring solar radiation) at the surface. However, care should be taken when comparing model parameters with observations, because observations are often local to a particular point in space and time, rather than representing averages over a model grid box. 
    
    This parameter is accumulated over a particular time period which depends on the data extracted. The units are joules per square metre (J m-2). To convert to watts per square metre (W m-2), the accumulated values should be divided by the accumulation period expressed in seconds. The ECMWF convention for vertical fluxes is positive downwards. &0.000000 \\
    \midrule
    Solar radiation (clear sky)    & Jm$^{-2}$ & Clear-sky downward shortwave radiation flux at surface computed from the model radiation scheme. &0.000000 \\
    \midrule
    Solar radiation (top of atmosphere)    & Jm$^{-2}$ & This parameter is the incoming solar radiation (also known as shortwave radiation) minus the outgoing solar radiation at the top of the atmosphere. It is the amount of radiation passing through a horizontal plane. The incoming solar radiation is the amount received from the Sun. The outgoing solar radiation is the amount reflected and scattered by the Earth's atmosphere and surface.
    
    This parameter is accumulated over a particular time period which depends on the data extracted. The units are joules per square metre (J m-2). To convert to watts per square metre (W m-2), the accumulated values should be divided by the accumulation period expressed in seconds.
    
    The ECMWF convention for vertical fluxes is positive downwards &0.000000 \\
    \midrule
    Solar radiation (total sky)    & J m$^{-2}$  & This parameter is the amount of solar (shortwave) radiation reaching the surface of the Earth (both direct and diffuse) minus the amount reflected by the Earth's surface (which is governed by the albedo), assuming clear-sky (cloudless) conditions. It is the amount of radiation passing through a horizontal plane. Clear-sky radiation quantities are computed for exactly the same atmospheric conditions of temperature, humidity, ozone, trace gases and aerosol as the corresponding total-sky quantities (clouds included), but assuming that the clouds are not there. Radiation from the Sun (solar, or shortwave, radiation) is partly reflected back to space by clouds and particles in the atmosphere (aerosols) and some of it is absorbed. The rest is incident on the Earth's surface, where some of it is reflected. The difference between downward and reflected solar radiation is the surface net solar radiation. This parameter is accumulated over a particular time period which depends on the data extracted. For the reanalysis, the accumulation period is over the 1 hour ending at the validity date and time. For the ensemble members, ensemble mean and ensemble spread, the accumulation period is over the 3 hours ending at the validity date and time. The units are joules per square metre (J m-2 ). To convert to watts per square metre (W m-2 ), the accumulated values should be divided by the accumulation period expressed in seconds. The ECMWF convention for vertical fluxes is positive downwards.  & 0.000000\\
    \midrule
    Solar radiation (top of atmosphere) (clear sky)    & J m$^{-2}$ & This parameter is the incoming solar radiation (also known as shortwave radiation) minus the outgoing solar radiation at the top of the atmosphere, assuming clear-sky (cloudless) conditions. It is the amount of radiation passing through a horizontal plane. The incoming solar radiation is the amount received from the Sun. The outgoing solar radiation is the amount reflected and scattered by the Earth's atmosphere and surface, assuming clear-sky (cloudless) conditions. Clear-sky radiation quantities are computed for exactly the same atmospheric conditions of temperature, humidity, ozone, trace gases and aerosol as the total-sky (clouds included) quantities, but assuming that the clouds are not there. This parameter is accumulated over a particular time period which depends on the data extracted. For the reanalysis, the accumulation period is over the 1 hour ending at the validity date and time. For the ensemble members, ensemble mean and ensemble spread, the accumulation period is over the 3 hours ending at the validity date and time. The units are joules per square metre (J m-2 ). To convert to watts per square metre (W m-2 ), the accumulated values should be divided by the accumulation period expressed in seconds. The ECMWF convention for vertical fluxes is positive downwards.  & 0.000000\\
    \midrule
    Temperature    & K & This parameter is the temperature in the atmosphere. It has units of kelvin (K). Temperature measured in kelvin can be converted to degrees Celsius (°C) by subtracting 273.15.  &272.976929 \\
    \midrule
    Surface pressure      & Pa & This parameter is the pressure (force per unit area) of the atmosphere at the surface of land, sea and inland water. It is a measure of the weight of all the air in a column vertically above a point on the Earth's surface. Surface pressure is often used in combination with temperature to calculate air density. The strong variation of pressure with altitude makes it difficult to see the low and high pressure weather systems over mountainous areas, so mean sea level pressure, rather than surface pressure, is normally used for this purpose. The units of this parameter are Pascals (Pa). Surface pressure is often measured in hPa and sometimes is presented in the old units of millibars, mb (1 hPa = 1 mb= 100 Pa). & 99115.242188\\
    \midrule
    Thermal radiation     & Jm$^{-2}$ & This parameter is the amount of thermal (also known as longwave or terrestrial) radiation emitted by the atmosphere and clouds that reaches a horizontal plane at the surface of the Earth. The surface of the Earth emits thermal radiation, some of which is absorbed by the atmosphere and clouds. The atmosphere and clouds likewise emit thermal radiation in all directions, some of which reaches the surface (represented by this parameter). This parameter is accumulated over a particular time period which depends on the data extracted. The units are joules per square metre (J m-2). To convert to watts per square metre (W m-2), the accumulated values should be divided by the accumulation period expressed in seconds.  & 845375.562500 \\
    \midrule
    Thermal radiation (clear sky)    & Jm$^{-2}$ & Clear-sky downward longwave radiation flux at surface computed from the model radiation scheme. &849147.312500 \\
    \midrule
    Thermal radiation (top of atmosphere)    & J m$^{-2}$ & The thermal (also known as terrestrial or longwave) radiation emitted to space at the top of the atmosphere is commonly known as the Outgoing Longwave Radiation (OLR). The top net thermal radiation (this parameter) is equal to the negative of OLR. This parameter is accumulated over a particular time period which depends on the data extracted. For the reanalysis, the accumulation period is over the 1 hour ending at the validity date and time. For the ensemble members, ensemble mean and ensemble spread, the accumulation period is over the 3 hours ending at the validity date and time. The units are joules per square metre (J m-2 ). To convert to watts per square metre (W m-2 ), the accumulated values should be divided by the accumulation period expressed in seconds. The ECMWF convention for vertical fluxes is positive downwards.  &-854573.250000 \\
    \midrule
    Thermal radiation (top of atmosphere) (clear sky)     & J m$^{-2}$ &This parameter is the thermal (also known as terrestrial or longwave) radiation emitted to space at the top of the atmosphere, assuming clear-sky (cloudless) conditions. It is the amount passing through a horizontal plane. Note that the ECMWF convention for vertical fluxes is positive downwards, so a flux from the atmosphere to space will be negative. Clear-sky radiation quantities are computed for exactly the same atmospheric conditions of temperature, humidity, ozone, trace gases and aerosol as total-sky quantities (clouds included), but assuming that the clouds are not there. The thermal radiation emitted to space at the top of the atmosphere is commonly known as the Outgoing Longwave Radiation (OLR) (i.e., taking a flux from the atmosphere to space as positive). Note that OLR is typically shown in units of watts per square metre (W m-2 ). This parameter is accumulated over a particular time period which depends on the data extracted. For the reanalysis, the accumulation period is over the 1 hour ending at the validity date and time. For the ensemble members, ensemble mean and ensemble spread, the accumulation period is over the 3 hours ending at the validity date and time. The units are joules per square metre (J m-2 ). To convert to watts per square metre (W m-2 ), the accumulated values should be divided by the accumulation period expressed in seconds.  &-853921.937500 \\
    \midrule
    Total cloud cover      & Dimensionless  & This parameter is the proportion of a grid box covered by cloud. Total cloud cover is a single level field calculated from the cloud occurring at different model levels through the atmosphere. Assumptions are made about the degree of overlap/randomness between clouds at different heights. Cloud fractions vary from 0 to 1.  &0.224129 \\
    \midrule
    Total precipitation     & m  &This parameter is the accumulated liquid and frozen water, comprising rain and snow, that falls to the Earth's surface. It is the sum of large-scale precipitation and convective precipitation. Large-scale precipitation is generated by the cloud scheme in the ECMWF Integrated Forecasting System (IFS). The cloud scheme represents the formation and dissipation of clouds and large-scale precipitation due to changes in atmospheric quantities (such as pressure, temperature and moisture) predicted directly by the IFS at spatial scales of the grid box or larger. Convective precipitation is generated by the convection scheme in the IFS, which represents convection at spatial scales smaller than the grid box. This parameter does not include fog, dew or the precipitation that evaporates in the atmosphere before it lands at the surface of the Earth. This parameter is accumulated over a particular time period which depends on the data extracted. For the reanalysis, the accumulation period is over 1 hour, ending at the validity date and time. For the ensemble members, ensemble mean and ensemble spread, the accumulation period is over the 3 hours ending at the validity date and time. The units of this parameter are depth in metres of water equivalent. It is the depth the water would have if it were spread evenly over the grid box. Care should be taken when comparing model parameters with observations, because observations are often local to a particular point in space and time, rather than representing averages over a model grid box.  &0.000000 \\
    \bottomrule
    \label{tab:features}
\end{longtable}

\end{document}